%% file: egpaper_final.tex
\documentclass[10pt,twocolumn,letterpaper]{article}
\usepackage{titling}
\usepackage{iccv}
\usepackage{times}
\usepackage{epsfig}
\usepackage{graphicx}
\usepackage{amsmath}
\usepackage{amssymb}
\usepackage{pdfpages}

\usepackage[export]{adjustbox}
\usepackage{color}
\usepackage{url}
\usepackage{gensymb}
\usepackage{multirow}
\usepackage{dcolumn}
  
\makeatletter
\renewcommand*\env@matrix[1][\arraystretch]{%
  \edef\arraystretch{#1}%
  \hskip -\arraycolsep
  \let\@ifnextchar\new@ifnextchar
  \array{*\c@MaxMatrixCols c}}
\makeatother

\usepackage[pagebackref=true,breaklinks=true,letterpaper=true,colorlinks,bookmarks=false,hypertexnames=false]{hyperref}
\usepackage[capitalize,nameinlink]{cleveref}
\iccvfinalcopy % *** Uncomment this line for the final submission

\def\iccvPaperID{3435} % *** Enter the ICCV Paper ID here

\begin{document}

%%%%%%%%% TITLE
\title{Calibration Wizard: A Guidance System for Camera Calibration\\Based on Modelling Geometric and Corner Uncertainty}
%\title{Calibration Wizard: A Guidance System for Camera Calibration by Modelling Geometry Uncertainty}

\author{Songyou Peng\thanks{Most work was done while he was an intern at INRIA.}\\
ETH Zurich\\
{\tt\small songyou.peng@inf.ethz.ch}
\and
Peter Sturm\\
INRIA Grenoble -- Rh\^{o}ne-Alpes \\
{\tt\small peter.sturm@inria.fr}
}

\maketitle
%\thispagestyle{empty}

%%%%%%%%% ABSTRACT
\begin{abstract}
   It is well known that the accuracy of a calibration depends strongly on the choice of camera poses from which images of a calibration object are acquired. We present a system -- Calibration Wizard -- that interactively guides a user towards taking optimal calibration images. For each new image to be taken, the system computes, from all previously acquired images, the pose that leads to the globally maximum reduction of expected uncertainty on intrinsic parameters and then guides the user towards that pose. We also show how to incorporate uncertainty in corner point position in a novel principled manner, for both, calibration and computation of the next best pose. Synthetic and real-world experiments are performed to demonstrate the effectiveness of \href{https://github.com/pengsongyou/CalibrationWizard}{Calibration Wizard}.
\end{abstract}

%%%%%%%%% BODY TEXT
%--------------------------------------------------------------------------
\section{Introduction}
\label{sec:intro}

\begin{figure*}[!ht]
  \centering
  \begin{tabular}{ccc}
    \includegraphics[width = 0.27\linewidth]{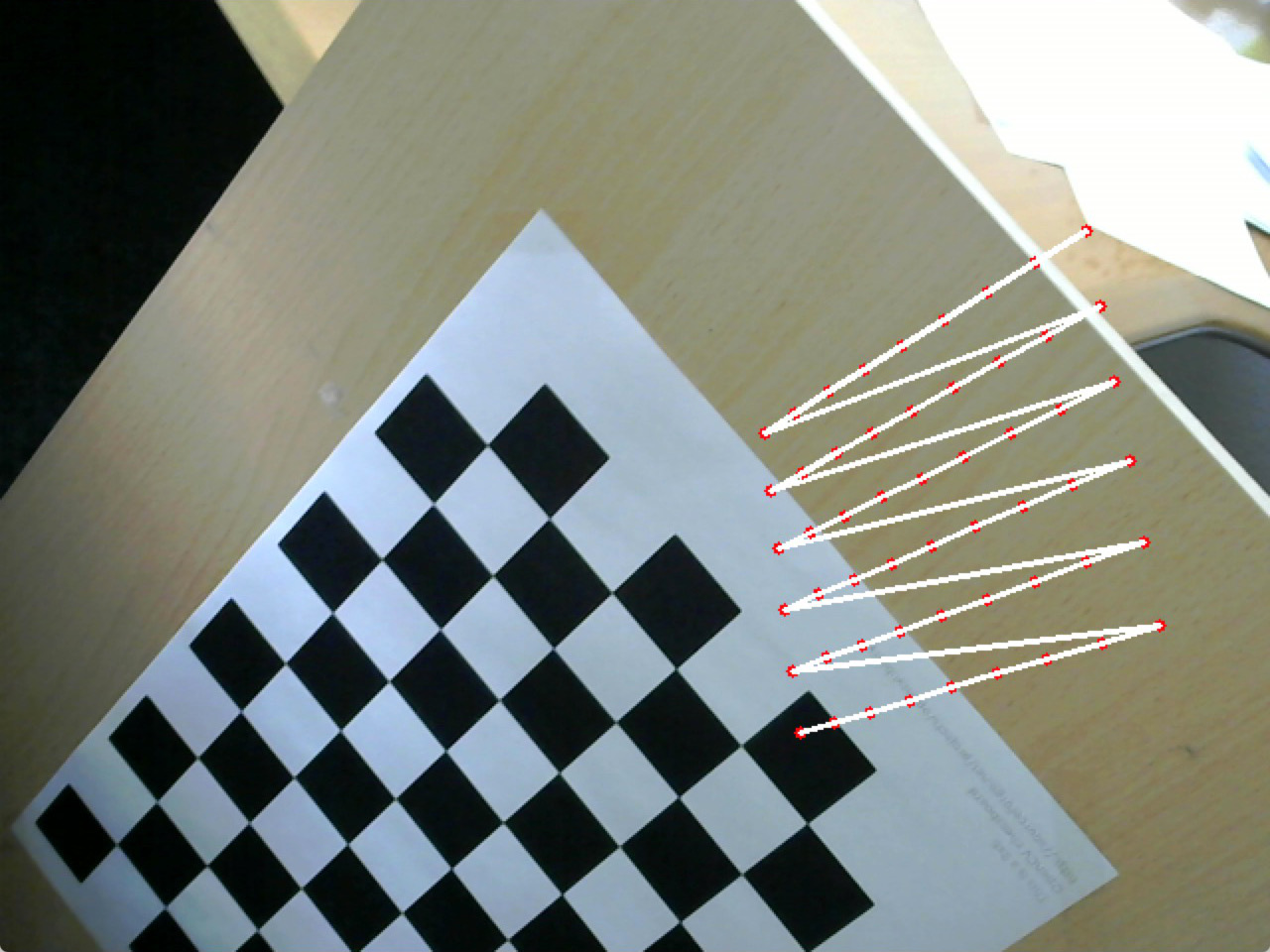}  &
    \includegraphics[width = 0.27\linewidth]{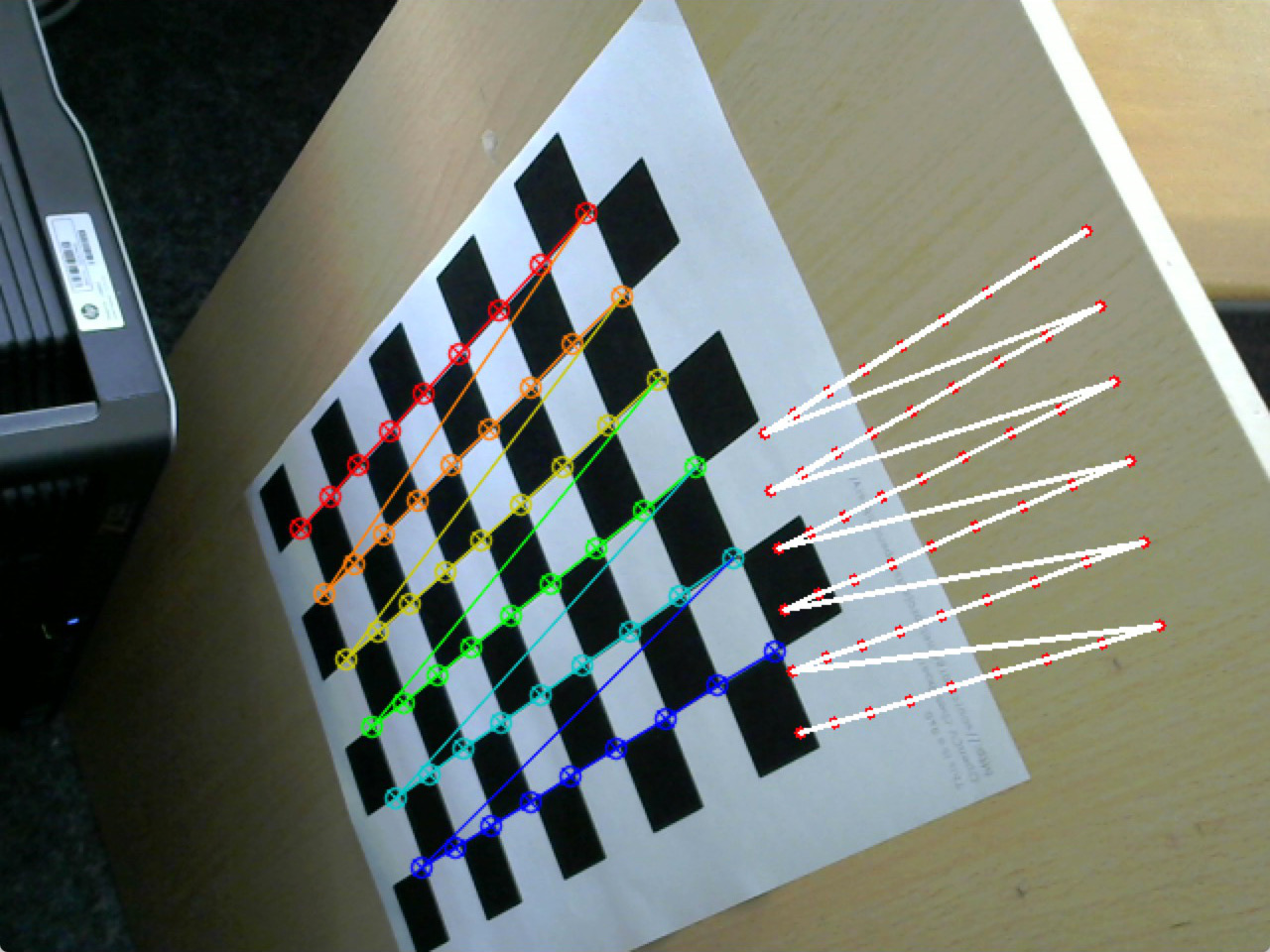} & 
    \includegraphics[width = 0.27\linewidth]{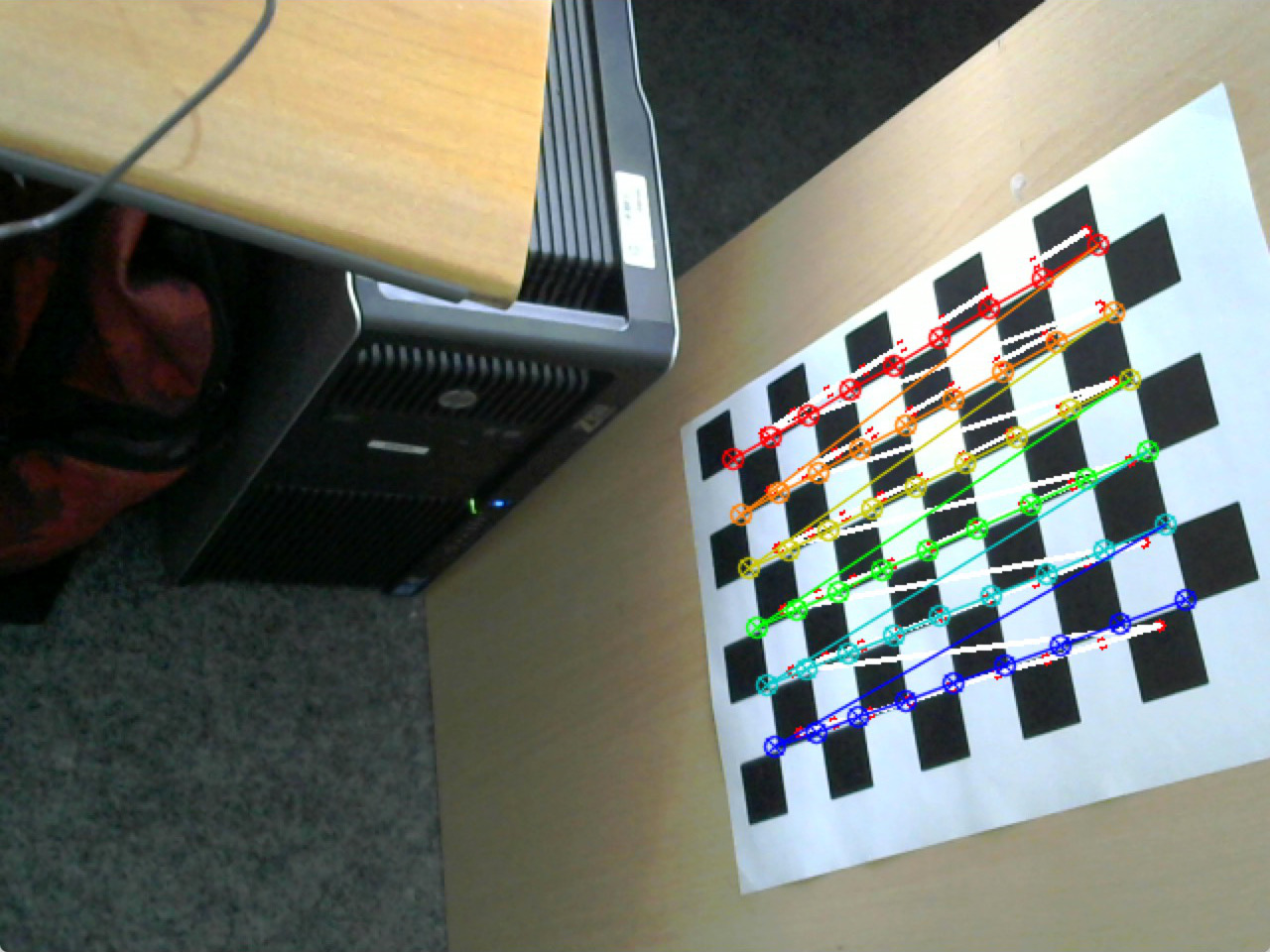}  \\         
    
    {\small (a)} &  {\small (b)} & {\small (c)} 
  \end{tabular}
\caption{Illustration of the guidance process. (a) Calibration Wizard proposes the next best pose based on the previous calibration results. (b) The camera should be moved towards the proposed target pose. (c) When the camera is close enough to the suggested pose, the system acquires an image and then proposes a next pose. Demo code:~\url{https://github.com/pengsongyou/CalibrationWizard}.}
\label{fig:wizard-process}
\end{figure*}

Camera calibration is a prerequisite to many methods and applications in computer vision and photogrammetry, in particular for most problems where 3D reconstruction or motion estimation is required.
In this paper we adopt the popular usage of planar calibration objects, as introduced\footnote{Planar targets were used before, but essentially in combination with motion stages, in order to effectively generate 3D targets \cite{tsai1987versatile,lenz1988techniques,wei1993complete,li1994camera}} by \cite{sturm1999plane,zhang2000flexible} and made available through OpenCV~\cite{opencv} and Matlab~\cite{bouguet2000matlab}, even though our approach can be directly applied to 3D calibration objects too.

It is well known that when using planar calibration targets, the accuracy of the resulting calibration depends strongly on the poses used to acquire images.
From the theoretical study of degenerate sets of camera poses in \cite{sturm1999plane,zhang2000flexible}, it follows for example intuitively that it is important to vary the orientation (rotation) of the camera as much as possible during the acquisition process.
It is also widely known to practitioners that a satisfactory calibration requires images such that the target successively covers the entire image area, otherwise the estimation of radial distortion and other parameters usually remains suboptimal.
We have observed that inexperienced users usually do not take calibration images that lead to a sufficiently accurate calibration.

Several efforts have been done in the past to guide users in placing the camera.
In photogrammetry for instance, the so-called network design problem was addressed, through an off-line process: how to place a given number of cameras such as to obtain an as accurate as possible 3D reconstruction of an object of assumed proportions \cite{Mason1995a,Olague2002a}.
Optimal camera poses for camera calibration have been computed in~\cite{ricolfe2011camera}, however only for constrained camera motions and especially, only for the linear approach of \cite{zhang2000flexible}, whereas we consider the non-linear optimization for calibration.
Also, these poses are difficult to realize, even for expert users.
The ROS~\cite{quigley2009ros} monocular camera calibration toolbox provides text instructions so that users can move the target accordingly.
More recently, some cameras, such as the ZED stereo system from StereoLabs, come with software that interactively guides the user to good poses during the calibration process; these poses are however pre-computed for the particular stereo system and this software cannot be used to calibrate other systems, especially monocular cameras.

In this paper we propose a system that guides the user through a simple graphical user interface (GUI) in order to move the camera to poses that are optimal for calibrating a camera. Optimality is considered for the bundle adjustment type non-linear optimization formulation for calibration. For each new image to be acquired, the system computes the optimal pose, i.e.\ which adds most new information on intrinsic parameters, in addition to that provided by the already acquired images.
% The most closely related work we are aware of is AprilCal~\cite{richardson2013aprilcal}. It also suggests a next best pose to the user. However, suggestions are selected from a fixed dataset of 60 pre-defined poses, which may not be enough for various camera models or calibration targets.
The most closely related works we are aware of are \cite{richardson2013aprilcal,rojtberg2018efficient}. They also suggest next best poses to the user. However, unlike ours, they are both strategy-based methods, where suggestions are selected from a fixed dataset of pre-defined poses, which may not be enough for various camera models or calibration targets.
In our approach, each new suggested pose results from a global optimization step.
Furthermore, we propose a novel method for incorporating the uncertainty of corner point positions rigorously throughout the entire pipeline, for calibration but also next best pose computation.
Our approach is not specific to any camera model; in principle, any monocular camera model can be plugged into it, although tests with very wide field of view cameras need to be done.

The paper is organized as follows. Section~\ref{sec:method} describes the theory and mathematical details of Calibration Wizard. Section~\ref{sec:autocorrelation} shows how to incorporate corner point uncertainty in the process.
% An intuitive way of displaying calibration uncertainty on the pixel level, is presented in Section~\ref{sec:uncertainty-map}.
Experiments are reported in Section~\ref{sec:results}, followed by conclusions in Section~\ref{sec:conclusions}.

%--------------------------------------------------------------------------
\section{Methodology}
\label{sec:method}

Our goal is to provide an interactive guidance for the acquisition of good images for camera calibration.
An initial calibration is carried out from a few (typically 3) freely taken images of a calibration target.
The system then guides the user to successive next best poses, through a simple GUI, cf.\ Fig~\ref{fig:wizard-process}.
The underlying computations are explained in the following subsections.

%%%%%%%%%%%%%%%%%%%%%%%%%%%%%%%%%%%%%%%%%%%%%%%%%%%
\subsection{Calibration formulation}
\label{method-formulation}

Let the camera be modeled by two \textbf{local projection functions} $x=q_x(\Theta, S), y=q_y(\Theta, S)$ that map a 3D point $S$ given in the local camera coordinate system, to the image coordinates $x$ and $y$.
These functions depend on intrinsic parameters $\Theta$ (assumed constant across all images).
For example, the standard 3-parameter pinhole model consisting of a focal length $f$ and principal point $(u,v)$, i.e.\ with $\Theta=(f, u, v)^\top$, has the following local projection functions (a full model with radial distortion is handled in supplementary material, section 3):
\begin{equation}
q_x(\Theta,S) = u + f \frac{S_1}{S_3}
\enspace\enspace\enspace
q_y(\Theta,S) = v + f \frac{S_2}{S_3}
\label{eq:pinhole}
\end{equation}

Let camera pose be given by a 6-vector $\Pi$ of extrinsic parameters.
We use the representation $\Pi=(t_1,t_2,t_3,\alpha,\beta,\gamma)^\top$, where $t=(t_1,t_2,t_3)^\top$ is a translation vector and the 3 angles define a rotation matrix as the product of rotations about the 3 coordinate axes: $R = R_z(\gamma) R_y(\beta) R_x(\alpha)$.
$R$ and $t$ map 3D points $Q$ from the world coordinate system to the local camera coordinate system according to:
\begin{equation}
	S = RQ + t
\end{equation}
Other parameterization may be used too, e.g.\ quaternions.

A camera with pose $\Pi$ is thus described by two \textbf{global projection functions} $p_x$ and $p_y$:
\begin{align}
	p_x(\Theta, \Pi, Q) &= q_x(\Theta, RQ+t) = q_x(\Theta, S)\\
	p_y(\Theta, \Pi, Q) &= q_y(\Theta, RQ+t) = q_y(\Theta, S)
\end{align}
Since a planar calibration target is used, the 3D points $Q$ are pre-defined and their corresponding $Z$ coordinates $Q_3$ are set to $0$.
We now consider $m$ images of a target consisting of $n$ calibration points.
Inputs to the calibration are thus the image points $(x_{ij}, y_{ij})$ for $i=1\cdots m$ and $j=1\cdots n$, which are detected by any corner detector, e.g.\ the OpenCV \verb+findChessboardCorners+ function~\cite{opencv}.
For ease of explanation, we assume here that all points are visible in all images, although this is not required in the implementation.

Optimal calibration requires a non-linear simultaneous optimization of all intrinsic and extrinsic parameters (bundle adjustment).
This comes down to the minimization of the geometric reprojection error \cite{hartley2003multiple}:
\begin{equation}
   \min_{\Theta, \{\Pi_i\}} \sum_{i,j} \left( x_{ij} - p_x(\Theta, \Pi_i, Q_j) \right)^2 + \left( y_{ij} - p_y(\Theta, \Pi_i, Q_j) \right)^2
   \label{eq:cost}
\end{equation}
Usually, local non-linear least square optimizers are used, such as Levenberg-Marquardt.
Our system is independent of the optimizer used; all it requires is the computation, at the found solution, of the partial derivatives of \eqref{eq:cost}, see next.

%%%%%%%%%%%%%%%%%%%%%%%%%%%%%%%%%%%%%%%%%%%%%%%%%%%
\subsection{Computation of next best pose}
\label{ss:next_pose}

We suppose that we have already acquired $m$ images and estimated intrinsic parameters and poses from these, by solving \eqref{eq:cost}.
The goal now is to compute the next best pose; the objective is to reduce, as much as possible, the expected uncertainty on the estimated intrinsic parameters.

Let us consider the Jacobian matrix $J$ of the cost function~\eqref{eq:cost}, evaluated at the estimated parameters.
$J$ contains the partial derivatives of the cost function's residuals, i.e.\ of terms
$\hat{x}_{ij} = x_{ij} - p_x(\Theta, \Pi_i, Q_j)$ and $\hat{y}_{ij} = y_{ij} - p_y(\Theta, \Pi_i, Q_j)$.
$J$ contains one row per residual.
Its columns are usually arranged in groups, such that the first group contains the partial derivatives
with respect to the intrinsic parameters $\Theta$, and subsequent groups of columns, the derivatives
relative to extrinsic parameters of the successive images.
The highly sparse form of $J$ is thus as follows (we assume here that there are $k$ intrinsic parameters):
\begin{equation}
\hspace{-0.7em}
J = \begin{pmatrix}
A_1 & B_1 & 0 & \cdots & 0 \\
A_2 & 0 & B_2 & \cdots & 0 \\
\vdots & \vdots & \vdots & \ddots & \vdots \\
A_m & 0 & 0 & \cdots & B_m
\end{pmatrix}
% \enspace
% A_i = \begin{pmatrix}
% \frac{\partial \hat{x}_{i1}}{\Theta_1} & \cdots & \frac{\partial \hat{x}_{i1}}{\Theta_k} \\
% \frac{\partial \hat{y}_{i1}}{\Theta_1} & \cdots & \frac{\partial \hat{y}_{i1}}{\Theta_k} \\
% \vdots & \ddots & \vdots \\
% \frac{\partial \hat{x}_{in}}{\Theta_1} & \cdots & \frac{\partial \hat{x}_{in}}{\Theta_k} \\
% \frac{\partial \hat{y}_{in}}{\Theta_1} & \cdots & \frac{\partial \hat{y}_{in}}{\Theta_k}
% \end{pmatrix}
% \enspace
% B_i = \begin{pmatrix}
% \frac{\partial \hat{x}_{i1}}{\Pi_{i,1}} & \cdots & \frac{\partial \hat{x}_{i1}}{\Pi_{i,6}} \\
% \frac{\partial \hat{y}_{i1}}{\Pi_{i,1}} & \cdots & \frac{\partial \hat{y}_{i1}}{\Pi_{i,6}} \\
% \vdots & \ddots & \vdots \\
% \frac{\partial \hat{x}_{in}}{\Pi_{i,1}} & \cdots & \frac{\partial \hat{x}_{in}}{\Pi_{i,6}} \\
% \frac{\partial \hat{y}_{in}}{\Pi_{i,1}} & \cdots & \frac{\partial \hat{y}_{in}}{\Pi_{i,6}}
% \end{pmatrix}
\label{eq:jacob2}
\end{equation}
\begin{equation*}
    A_i = \begin{pmatrix}
\frac{\partial \hat{x}_{i1}}{\Theta_1} & \cdots & \frac{\partial \hat{x}_{i1}}{\Theta_k} \\
\frac{\partial \hat{y}_{i1}}{\Theta_1} & \cdots & \frac{\partial \hat{y}_{i1}}{\Theta_k} \\
\vdots & \ddots & \vdots \\
\frac{\partial \hat{x}_{in}}{\Theta_1} & \cdots & \frac{\partial \hat{x}_{in}}{\Theta_k} \\
\frac{\partial \hat{y}_{in}}{\Theta_1} & \cdots & \frac{\partial \hat{y}_{in}}{\Theta_k}
\end{pmatrix}
\enspace
B_i = \begin{pmatrix}
\frac{\partial \hat{x}_{i1}}{\Pi_{i,1}} & \cdots & \frac{\partial \hat{x}_{i1}}{\Pi_{i,6}} \\
\frac{\partial \hat{y}_{i1}}{\Pi_{i,1}} & \cdots & \frac{\partial \hat{y}_{i1}}{\Pi_{i,6}} \\
\vdots & \ddots & \vdots \\
\frac{\partial \hat{x}_{in}}{\Pi_{i,1}} & \cdots & \frac{\partial \hat{x}_{in}}{\Pi_{i,6}} \\
\frac{\partial \hat{y}_{in}}{\Pi_{i,1}} & \cdots & \frac{\partial \hat{y}_{in}}{\Pi_{i,6}}
\end{pmatrix}
\end{equation*}
where $A_i$ are matrices of size $2n \times k$, containing the partial derivatives of residuals with respect to $k$ intrinsic parameters,
whereas $B_i$ are matrices of size $2n \times 6$, containing the partial derivatives with respect to extrinsic parameters.
Now, the so-called information matrix is computed as $J^\top J$:
\begin{equation}
J^\top J = \begin{pmatrix}
\sum\limits_i A_i^\top A_i & A_1^\top B_1 & A_2^\top B_2 & \cdots & A_m^\top B_m \\
B_1^\top A_1 & B_1^\top B_1 & 0 & \cdots & 0 \\
B_2^\top A_2 & 0 & B_2^\top B_2 & \cdots & 0 \\
\vdots & \vdots & \vdots & \ddots & \vdots \\
B_m^\top A_m & 0 & 0 & \cdots & B_m^\top B_m
\end{pmatrix}
\label{eq:infoMat}
\end{equation}
Like $J$, $J^\top J $ is highly sparse.
Importantly, its inverse $(J^\top J)^{-1}$ provides an estimation of the covariance matrix of the estimated intrinsic and extrinsic parameters.
%is much easier to compute than the inverse of a non-sparse matrix of the same size.
%While it may not be possible to compute it analytically, its numerical computation is
%quite efficient (see pages 605 and 606 of the Hartley-Zisserman book, 2003 edition).

For camera calibration, we are only interested in the covariance matrix of the intrinsic parameters, i.e.\ the upper-left $k\times k$ sub-matrix of $(J^\top J)^{-1}$.
Due to the special structure of $(J^\top J)^{-1}$, it can be computed efficiently.
Let
\begin{align*}
    U &= \sum_i A_i^\top A_i \\
    V &= \text{diag}\left(B_1^\top B_1, \cdots, B_m^\top B_m\right) \\
    W &= \begin{pmatrix} A_1^\top B_1 & \cdots & A_m^\top B_m \end{pmatrix}
\end{align*}
% \[
% U = \sum_i A_i^\top A_i, \enspace\enspace\enspace
% W = \begin{pmatrix} A_1^\top B_1 & \cdots & A_m^\top B_m \end{pmatrix}, \enspace\enspace\enspace
% V = \text{diag}\left(B_1^\top B_1, \cdots, B_m^\top B_m\right)
% \]
Then,
\begin{equation}
    J^\top J = 
        \begin{pmatrix}
            U & W\\ W^\top & V
        \end{pmatrix}
\end{equation}
and as described in~\cite{hartley2003multiple}, the upper-left sub-matrix of $(J^\top J)^{-1}$ is given by $\Sigma = (U-W V^{-1} W^\top)^{-1}$, i.e.\ the inverse of a $k\times k$ symmetric matrix.

Let us now return to our goal, determining the next best pose $\Pi_{m+1}$.
The outline of how to achieve this is as follows.
We extend the Jacobian matrix in Eq.~\eqref{eq:jacob2} with a part corresponding to an additional image, whose pose is
parameterized by $\Pi_{m+1}$.
The coefficients in $A_{m+1}$ and $B_{m+1}$, are thus functions of $\Pi_{m+1}$.
Naturally, $\Pi_{m+1}$ is also implicitly embedded in the intrinsic parameter's covariance matrix associated with this extended system.
To reduce the uncertainty of the calibration, we wish to determine $\Pi_{m+1}$ that makes $\Sigma$ as ``small'' as possible.
Inspired by~\cite{haner2015phd}, we choose to minimize the trace of this $k \times k$ matrix.
Since we wish to compute the next best pose within the entire 3D working space, we use a global optimization method.
Our experiments suggest that simulated annealing~\cite{van1987simulated} or ISRES~\cite{runarsson2005search} work well for this small optimization problem~\footnote{We did not consider the stopping criterion in the current version, but one could simply stop our method when the relative residual of the trace of the covariance matrix mentioned above is smaller than a threshold.}.
Especially the latter works in interactive time.

Note that the computation of the partial derivatives used to build the $A_i$ and $B_i$ matrices can be done very efficiently using the chain rule.
Further, computation of $\Sigma$ for different trials of $\Pi_{m+1}$ can also be done highly efficiently by appropriate pre-computations of the parts of matrices $U$ and $W V^{-1} W^\top$ that do not depend on $\Pi_{m+1}$. See more details in the supplementary material, section 2. 
% \footnote{Supplementary material contains more details about partial derivatives in $J$ and the efficient computation of $C$.}.
% \footnote{More details can be found in supplementary material.}.

%-------------------------------------------------------------------------
%%%%%%%%%%%%%%%%%%%%%%%%%%%%%%%%%%%%%%%%%%%%%%%%%%%
\section{Taking into Account the Uncertainty of Corner Points}
\label{sec:autocorrelation}

\begin{figure*}[t]\centering
{\renewcommand{\arraystretch}{0.4}
\begin{tabular}{cccccc}
\multirow{3}{*}[1.2cm]{\includegraphics[width = 0.33\linewidth]{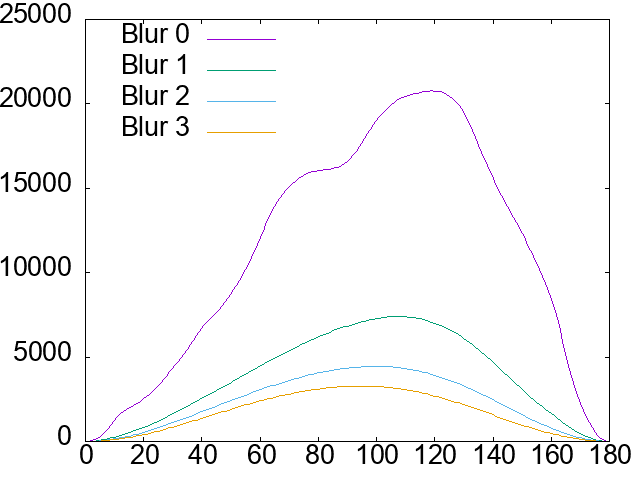}} &
\includegraphics[width = 0.07\linewidth, frame]{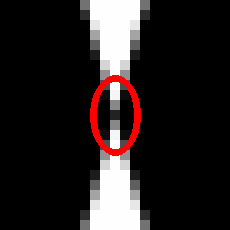}&
\includegraphics[width = 0.07\linewidth, frame]{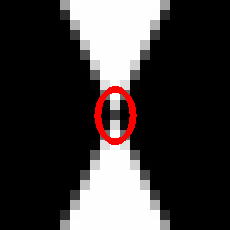}&
\includegraphics[width = 0.07\linewidth, frame]{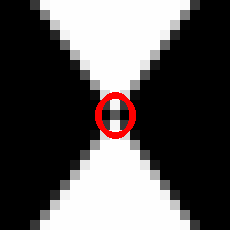}&
\includegraphics[width = 0.07\linewidth, frame]{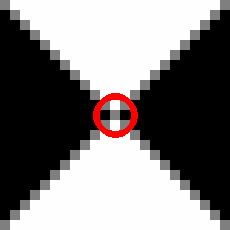}&
\includegraphics[width = 0.07\linewidth, frame]{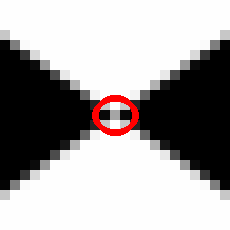} \\
& \includegraphics[width = 0.07\linewidth, frame]{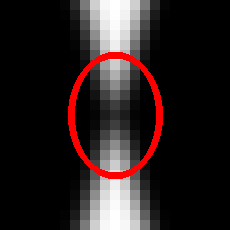}&
\includegraphics[width = 0.07\linewidth, frame]{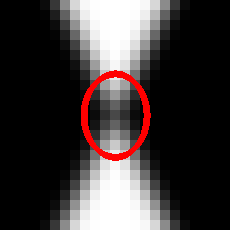}&
\includegraphics[width = 0.07\linewidth, frame]{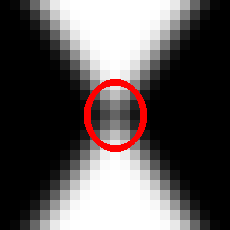}&
\includegraphics[width = 0.07\linewidth, frame]{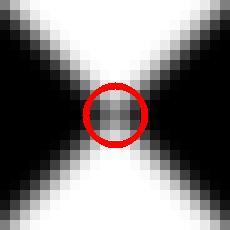}&
\includegraphics[width = 0.07\linewidth, frame]{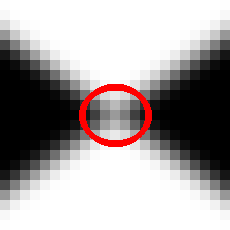} \\
& \includegraphics[width = 0.07\linewidth, frame]{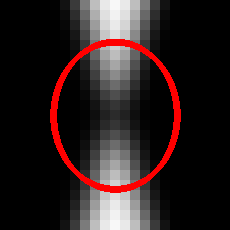}&
\includegraphics[width = 0.07\linewidth, frame]{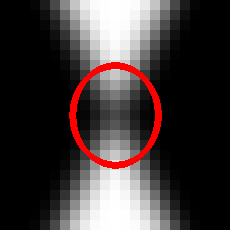}&
\includegraphics[width = 0.07\linewidth, frame]{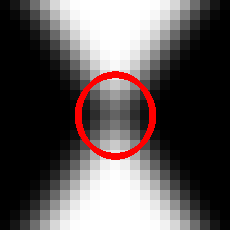}&
\includegraphics[width = 0.07\linewidth, frame]{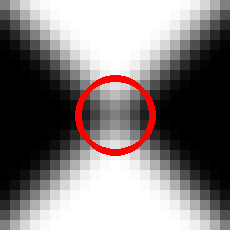}&
\includegraphics[width = 0.07\linewidth, frame]{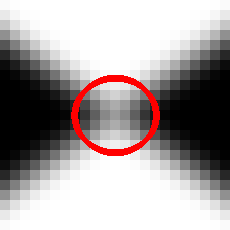}\\
& $30^\circ$& $50^\circ$& $70^\circ$& $90^\circ$& $110^\circ$
\end{tabular}}
\caption{Uncertainty of corner position as a function of opening angle and blur.
Left: plot of the first eigenvalue of the autocorrelation matrix, over opening angle and for different blur levels (Gaussian blur for $\sigma=0,1,2,3$).
Right: corners for different opening angles and blur levels ($\sigma=0,1,2$) and computed $95\%$ confidence level uncertainty ellipses (enlarged 10$\times$ for display).}
\label{fig:corner_uncertainty}
\end{figure*}

%\begin{figure*}[!ht]\centering
%\begin{tabular}{ccc}
%\includegraphics[width = 0.32\linewidth]{img/graph_blur_255.png} &
%\includegraphics[width = 0.32\linewidth]{img/graph_noise_255.png} &
%\includegraphics[width = 0.32\linewidth]{img/graph_noise_150.png}\\
%(a) & (b) & (c)\\
%\vspace{-1em}
%\end{tabular}
%\caption{(a) Second eigenvalue of auto-correlation matrix as a function of the corner's opening angle, for different blur levels.
%(b) same graph, for small blur ($\sigma=1$), with strong noise and without, for a greylevel range of 255. (c) same as center, but for a greylevel range of 150.}
%\label{fig:corner_uncertainty_graphs}
%\end{figure*}

So far, we have not used information on the uncertainty of corner point positions:
in Eq.~\eqref{eq:cost}, all residuals have the same weight (equal to $1$).
Ideally, in any geometric computer vision formulation, one should incorporate estimates of
uncertainty when available.
In the following, we explain this for our problem, from two aspects: first for the actual
calibration, i.e.\ the parameter estimation in Eq.~\eqref{eq:cost}.
Second, more originally, for computing the next best pose.

\subsection{Corner Uncertainty in Calibration}

Consider a corner point extracted in an image; the uncertainty of its position can be
estimated by computing the autocorrelation matrix $C$ for a window of a given size around
the point (see for instance \cite{harris1988a}).
Concretely, $C$ is an estimate of the inverse of the covariance matrix of the corner position.
Now, let $C_{ij}$ be the autocorrelation matrix for the $j$th corner in the $i$th image.
The $C_{ij}$ can be incorporated in the calibration process by inserting the block-diagonal
matrix composed by them, in the computation of the information matrix of Eq.~\eqref{eq:infoMat}:
\begin{equation}
H = J^\top \text{diag}(C_{11}, C_{12}, \cdots, C_{1n}, C_{21}, \cdots, C_{mn}) J
\label{eq.corrected_information_matrix}
\end{equation}

This uncertainty-corrected information matrix can then be used by Gauss-Newton or
Levenberg-Marquardt type optimizers for estimating the calibration (for other optimizers,
the autocorrelation matrix may have to be used differently) as well as for the quantification
of the uncertainty of the computed intrinsic parameters.

\subsection{Next Best Pose Computation}

The second usage of corner uncertainty concerns the computation of the next best pose.
In particular, for each hypothetical next pose that we examine, we can project the points of the calibration target to the image plane, using this hypothetical pose and the current estimates of intrinsic parameters.
This gives the expected positions of corner points, if an actual image were to be taken from that pose.
Importantly, what we would like to compute in addition, is the expected uncertainty of corner extraction, i.e.\ the uncertainty of the corner positions extracted in the expected image.
Or, equivalently, the autocorrelation matrices computed from the expected pixel neighborhoods around these corner points.
If we are able to do so, we can plug these into the estimation of the next best pose, by inserting the expected autocorrelation matrices in Eq.~\eqref{eq:infoMat} the same way as done in Eq.~\eqref{eq.corrected_information_matrix}.

Before explaining how to estimate expected autocorrelation matrices, we describe the benefits of this approach.
Indeed, we have found that without doing so, the next best pose is sometimes rather extreme,
with a strong grazing viewing angle relative to the calibration target.
This makes sense in terms of pure geometric information contributed by such a pose,
but is not appropriate in practice, since with extreme viewing angles image corners are
highly elongated: they may be difficult to extract in the actual image and also, their uncertainty is very large in one direction.
While this may be compensated by using images acquired from poses with approximately
perpendicular viewing directions one from another, it is desirable and indeed more principled to fully integrate
corner uncertainty right from the start in the computation of the next best pose.

Let us now explain how to compute expected autocorrelation matrices of corner points, for a hypothetical next pose.
This is based on a simple reasoning.
The overall shape of an image corner (in our case, a crossing in a checkerboard pattern), is entirely represented by the ``opening angle'' of the corner.
What we do is to precompute autocorrelation matrices for synthetic corners for the entire range of opening angles, cf.\ Fig.~\ref{fig:corner_uncertainty}: the top row on the right shows ideal corners generated by discretizing continuous black-and-white corners, for a few different opening angles.
For each of them, the autocorrelation matrix (cf.\ \cite{harris1988a}) was computed; as mentioned, its inverse is an estimate of the corner position's covariance matrix.
The figure shows the plots of $95\%$ uncertainty ellipses derived from these covariance matrices (enlarged 10 times, for better visibility).
Such ideal corners are of course not realistic, so we repeat the same process for images blurred by Gaussian kernels of different $\sigma$ (2nd and 3rd rows of the figure).
Naturally, blurrier corner images lead to smaller autocorrelation matrices and larger uncertainty ellipses.

One may note several things.
First, between $30^\circ$ and $90^\circ$ opening angles, the largest uncertainty differs by a factor of about $2$, see the uncertainty ellipses in Fig.~\ref{fig:corner_uncertainty}.
Second, intuitively, the uncertainty ellipse of a corner with opening angle $\alpha$ is the same as with opening angle $180^\circ - \alpha$, but turned by $90^\circ$ (cf.\ the 3rd and 5th columns in Fig.~\ref{fig:corner_uncertainty}, for $70^\circ$ and $110^\circ$).
Hence, the eigenvalues of the autocorrelation matrix of a corner with opening angle $\alpha$, are the same as that for $180^\circ - \alpha$, but they are ``swapped'' (associated with the respective opposite eigenvector).

The left part of Fig.~\ref{fig:corner_uncertainty} shows plots of the first eigenvalue (associated with eigenvector $(0,1)$) of the autocorrelation matrix $C$ as a function of opening angle.
Due to the above observation, the second eigenvalue (associated with eigenvector $(1,0)$) associated with opening angle $\alpha$ is simply given by the first eigenvalue associated with $180^\circ - \alpha$.
The graphs on the left of the figure confirm that increasing blur decreases the autocorrelation matrices eigenvalues.
Let us note that we also simulated Gaussian pixel noise on the corner images; even for larger than realistic noise levels, the impact on the results shown in Fig.~\ref{fig:corner_uncertainty}, was negligible.

Let us finally explain how to use these results.
First, we determine the average blur level in the already acquired images, from the strength of image gradients across edges, and then select the graph in Fig.~\ref{fig:corner_uncertainty} associated with the closest simulated blur (for more precision, one could also compute the graph for the actual detected blur level).
Let the function represented by the graph be $f(\alpha)$ -- one can represent it as a lookup table or fit a polynomial to the data of the graph (we did the latter).
This allows to compute the diagonal coefficients of the autocorrelation matrix from the opening angle,
as $f(\alpha)$ and $f(180^\circ - \alpha)$.
Second, so far we have only considered axis-aligned corners.
If we now consider a corner with opening angle $\alpha$, but that is rotated by an angle $\beta$, then
its autocorrelation matrix is nothing else than:
\begin{equation}
\footnotesize
C = \begin{pmatrix} \cos\beta & -\sin\beta \\ \sin\beta & \cos\beta \end{pmatrix}
\begin{pmatrix} f(\alpha) & 0 \\ 0 & f(180^\circ - \alpha) \end{pmatrix}
\begin{pmatrix} \cos\beta & \sin\beta \\ -\sin\beta & \cos\beta \end{pmatrix}
\end{equation}

We now have all that is needed to incorporate corner uncertainty in next best pose computation.
For each hypothetical pose we project, as shown above, all calibration points.
For each point (corner), using its neighbors, we can compute the opening angle $\alpha$
and rotation angle $\beta$ and thus, the expected autocorrelation matrix $C$.
It can then be inserted in the computation of the information matrix, like in Eq.~\eqref{eq.corrected_information_matrix}.

The effect of this approach on proposing the next best pose is to strike a balance between maximizing pure geometric ``strength'' of a pose (often achieved by strong tilting of the calibration pattern) and maximizing corner extraction accuracy (in fronto-parallel poses, corners are overall closest to exhibiting right angles, i.e.\ where their auto-correlation matrices have maximal trace).

\subsection{Possible Extensions}
\label{sec:autocorrelation.next}

So far we have described the basic idea for incorporating corner uncertainty.
The following extensions may be applied; we plan this for future work.
The values plotted in Fig.~\ref{fig:corner_uncertainty} are obtained for corners exhibiting the full range of 256 greylevels (black to white).
In real images, the range is of course smaller.
If the difference between largest and smallest greylevels is $x$, then the plotted values (left of Fig.~\ref{fig:corner_uncertainty}) are divided by $255^2/x^2$ (easy to prove but not shown due to lack of space).
In turn, the uncertainty ellipses are scaled up by a factor of $255/x$.
This should be taken into account when predicting auto-correlation matrices for the next pose.
The range of greylevels depends on various factors, such as distance to the camera and lighting conditions.
One can learn the relationship between pose and greylevel range for a given calibration setup as part of the calibration process and use it to predict the next best pose.

Similarly, one might predict expected blur per corner, based on a learned relationship between blur and distance to camera, e.g.\ by inferring the camera's depth of field during calibration. We observed that within the depth of field, image blur is linearly related to the distance to the camera.

Using these observations should allow to further improve the next best pose proposal, by achieving an even better compromise between geometric information of the pose and accuracy of image processing (here, corner extraction), both of which influence calibration accuracy.

%
%\textbf{Remarks.}
%The definition of the information matrix should also include the usage of a covariance
%matrix for the image points (matrix $R_i$ in equation (4.6) of Haner's PhD thesis).
%This is omitted here for simplicity.
%Most people assume the same covariance for all image points, in which case this
%covariance matrix can be left out in the definition of the information matrix.
%It would be good though, to include the possibility of using different covariance
%matrices for the image points, for usage of feature extraction methods that do provide
%such covariance matrices.

%--------------------------------------------------------------------------
\section{Results and Evaluation}
\label{sec:results}

Synthetic and real-world experiments are performed here to evaluate the effectiveness of our Calibration Wizard.
Note that in the optimization process, we ensure that all corner points should be within the image plane, otherwise the optimization loss is set to an extremely large value. 

\subsection{Synthetic evaluations}
To assess the proposed system, we simulate the process of camera calibration with pre-defined intrinsic parameters, with Matlab. Here we first briefly introduce the procedure of producing random checkerboard poses.

\textbf{Data preparation.}
First, $9\times 6$ target points are defined, with $Z$ components set to $0$.
Next, the 3D position of the virtual camera is created, with $X$ and $Y$ coordinates proportional to $Z$ within a plausible range.
Then the camera is first oriented such that its optical axis goes through the center of the target and finally, rotations about the three local coordinate axes by random angles between $-15^{\circ}$ and $15^{\circ}$ are applied.
Now, from the position, rotation matrix and the given intrinsic parameters, we can project the 3D target points to the image, and finally add zero-mean Gaussian noise with the same noise level to them.
Moreover, we ensure that all 54 points are located within the field of view of a $640\times 480$ image plane.

\begin{figure}[t]
  \centering
  \begin{tabular}{cc}
    \includegraphics[width = 0.48\linewidth]{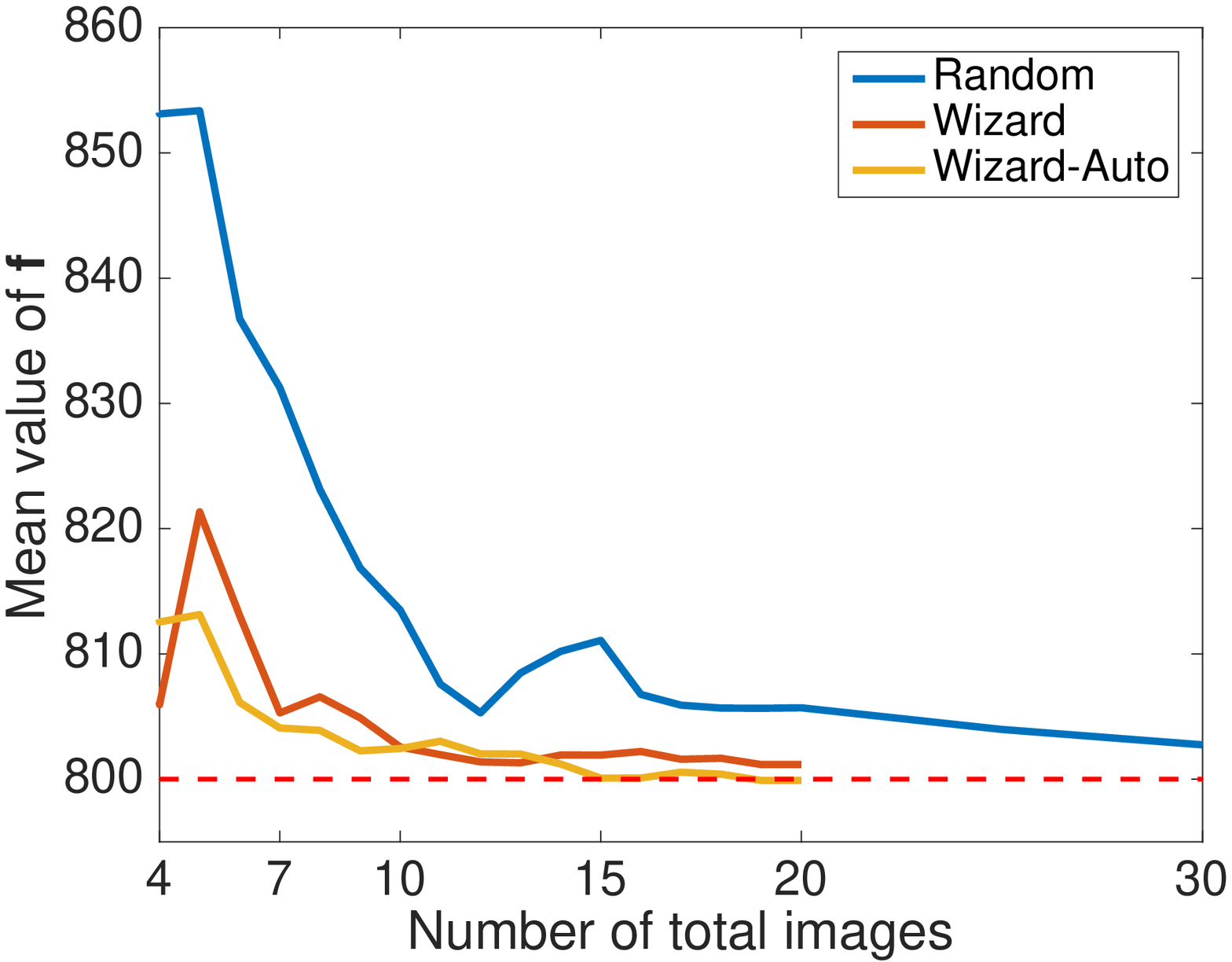}  &
    \includegraphics[width = 0.48\linewidth]{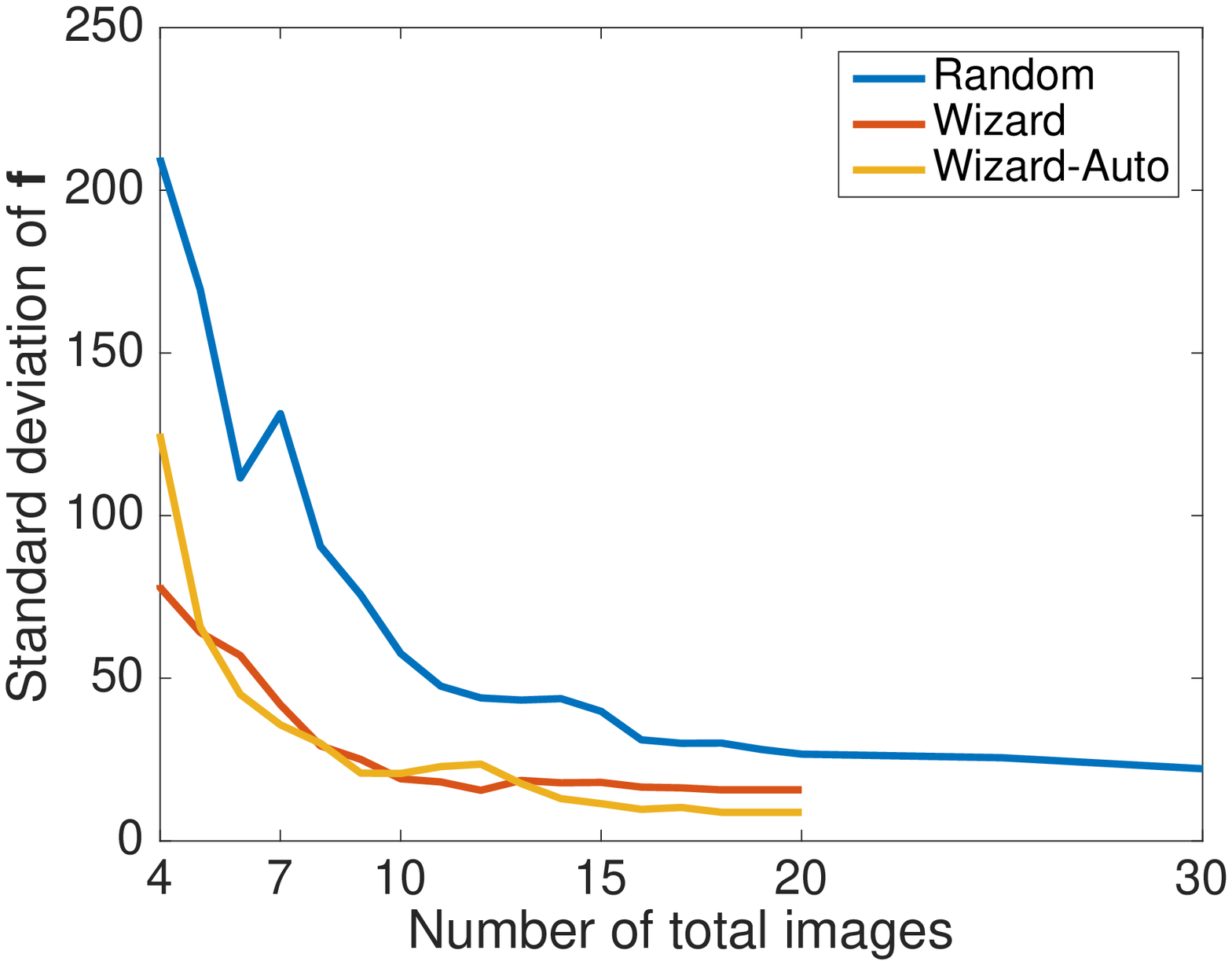}
%    \multicolumn{2}{c}{\includegraphics[width = 0.48\linewidth]{img/synCompZ}}\\
%    \multicolumn{2}{c}{\small (c)}
  \end{tabular}
\caption{Comparisons of the focal length estimated from three schemes on synthetic data: randomly taken images, calibration wizard and wizard using autocorrelation matrix. $f = 800, (u,v) = (320,240), k_1 = 0.01, k_2 = 0.1$. Initial calibration was done with $3$ random images. Left: Mean values of the estimated focal length, where the {\color{red} red} dashed line represents the ground truth $f=800$. Right: Standard deviations of the estimated focal length. Wizard images provide significantly more accurate and precise calibration results than random ones.}
\label{fig:syn-test1}
\end{figure}

\textbf{Evaluation of accuracy and precision.}
We primarily compare the calibration accuracy obtained from random images, with that from images acquired as proposed by our system, with and without taking into account the autocorrelation matrix explained in the previous section.
To this end, the experimental process is as follows: create 3 initial random images, based on which we have 3 paths to acquire the calibration results:
\begin{itemize}
    \setlength\itemsep{0.1em}
	\item Produce many other random images 
	\item Obtain 17 images proposed by the wizard, so $3 + 17 = 20$ images in total 
	\item Obtain 17 wizard images taking the autocorrelation matrix into account 
\end{itemize}
$100$ trials are performed for each experiment, hence we acquire $100$ samples of intrinsic parameters for each.
In the first test, we set $f = 800, (u,v) = (320,240)$ and radial distortion coefficients $ k_1 = 0.01, k_2 = 0.1$. 
Fig.~\ref{fig:syn-test1} illustrates the statistical results of the estimated focal length from $100$ trials. 
It can be easily noticed that the focal length acquired using our wizard is not only much closer to the ground truth, but also more concentrated (precise) than the estimation from pure random images.
For example, the estimated focal length acquired from only $3$ random + $4$ wizard images has outperformed the one from 20 random images.
Moreover, not shown in the graph: $3$ random + $17$ wizard images still give higher accuracy than $60$ random images, which directly demonstrates the usefulness of our approach. 

\begin{figure}[!t]
  \centering
  \begin{tabular}{cc}
    \includegraphics[width = 0.48\linewidth]{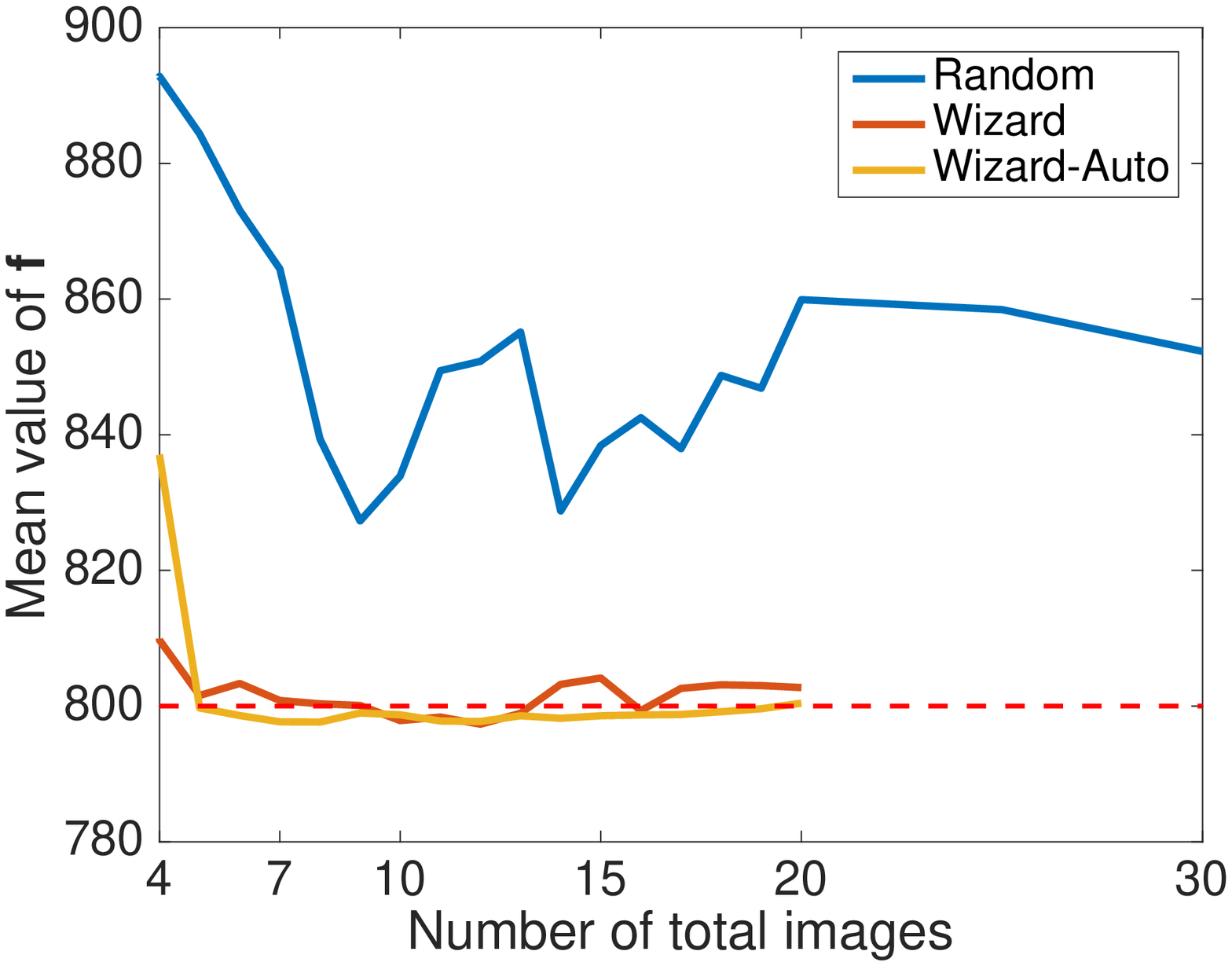}  &
    \includegraphics[width = 0.48\linewidth]{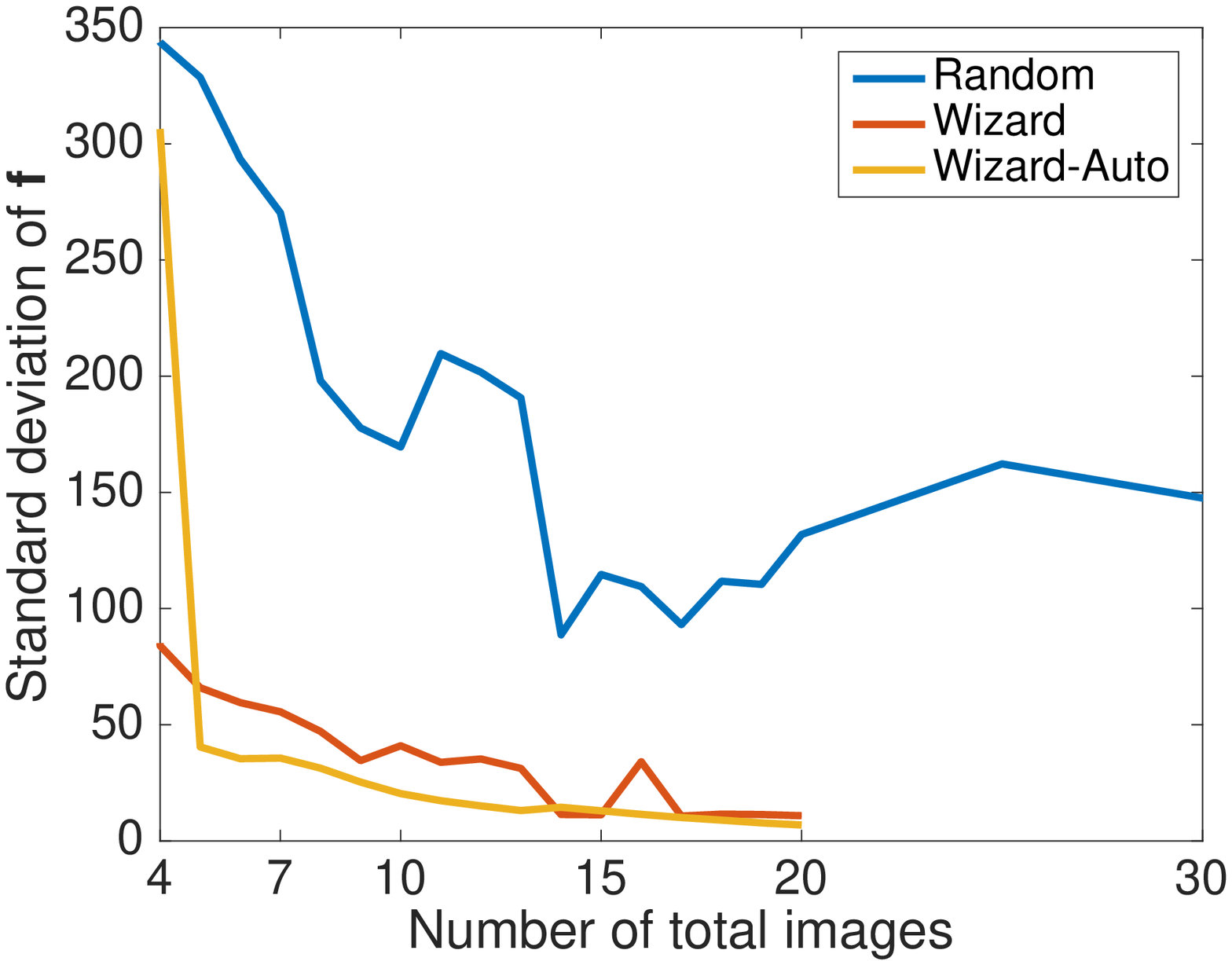}  \\         
    \includegraphics[width = 0.48\linewidth]{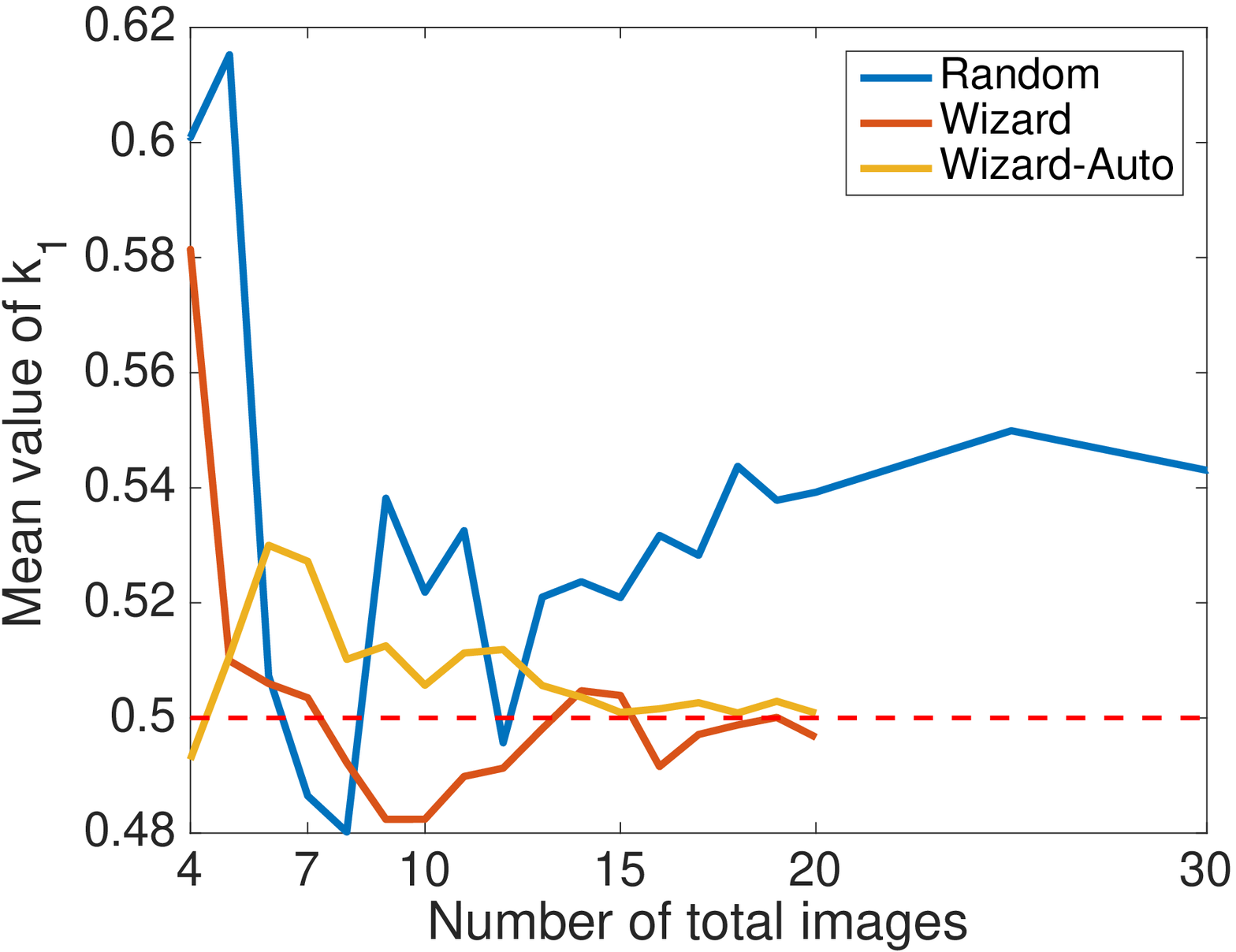}  &
    \includegraphics[width = 0.48\linewidth]{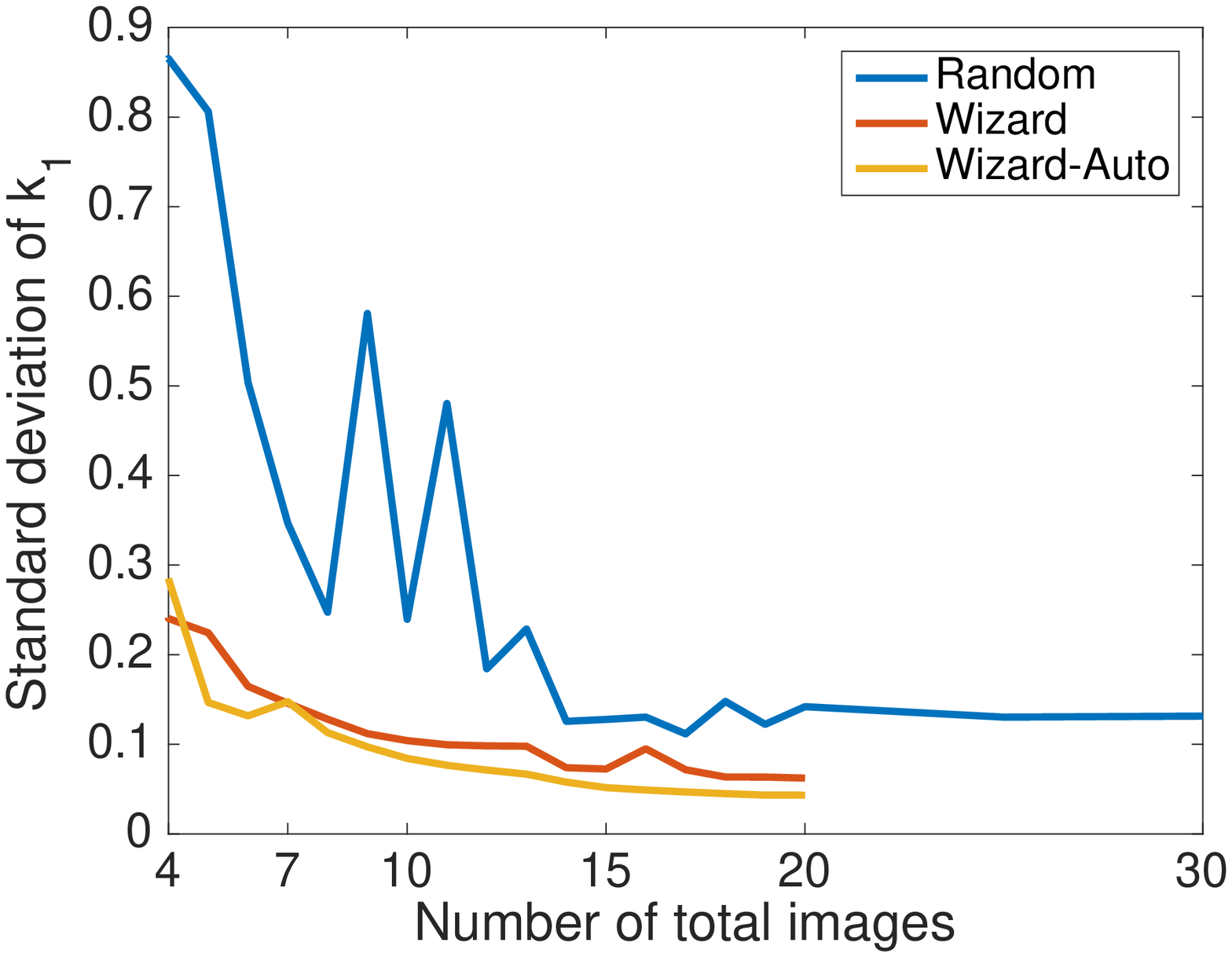}  \\             
    \includegraphics[width = 0.48\linewidth]{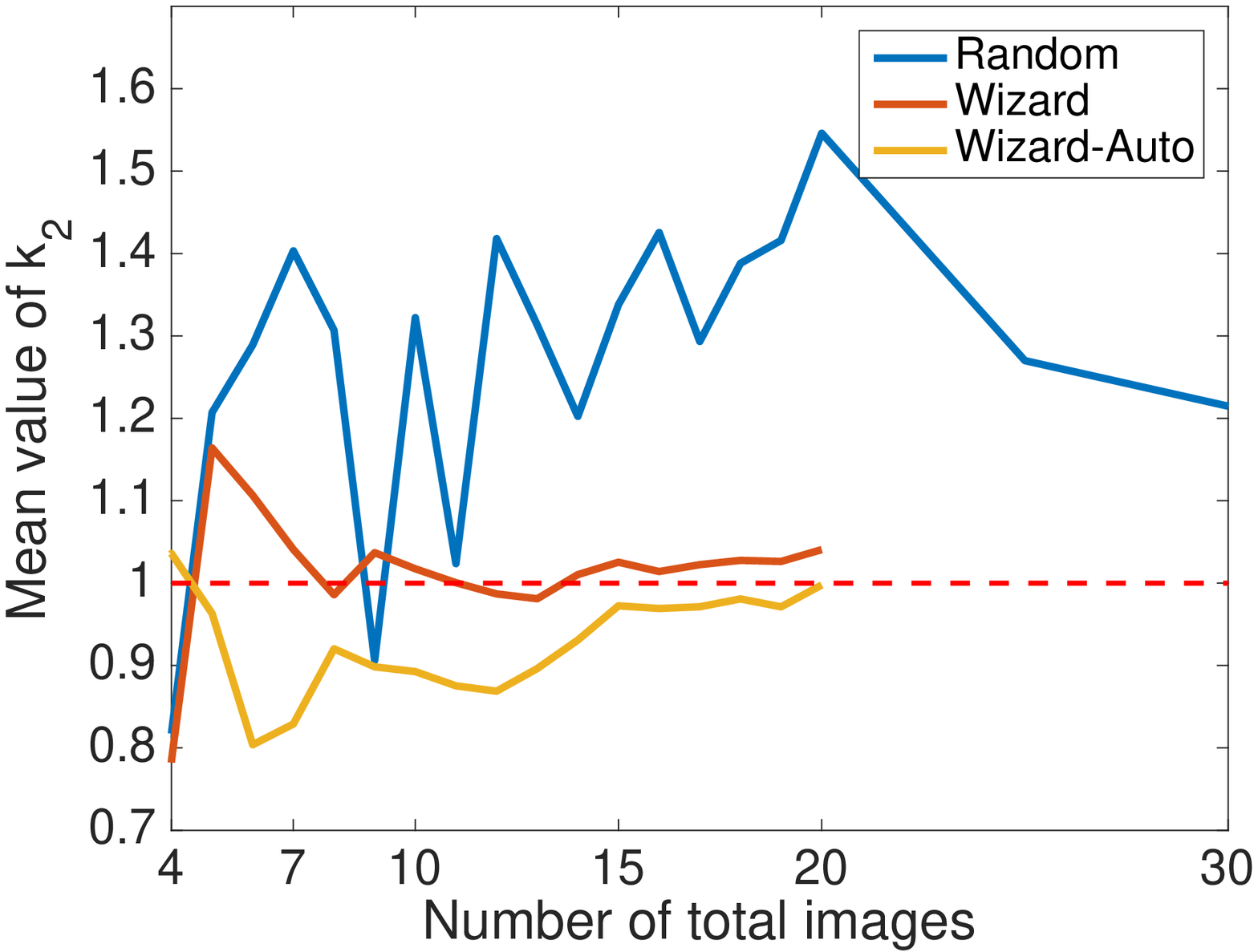}  &
    \includegraphics[width = 0.48\linewidth]{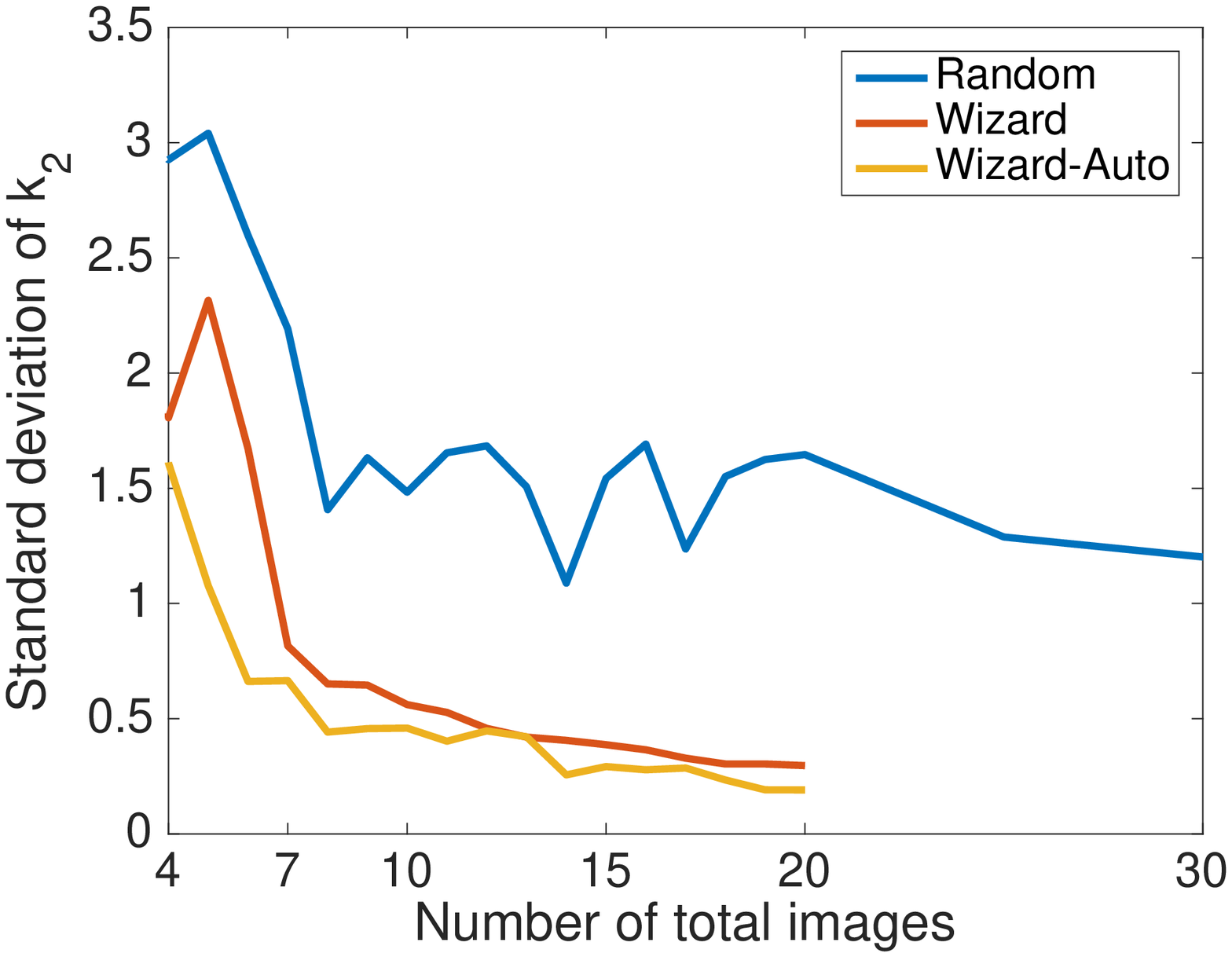}  \\                 
  \end{tabular}
\caption{Comparisons of the intrinsic parameters estimated from three schemes on synthetic data: randomly taken images, calibration wizard without and with autocorrelation matrix. $f = 800, (u,v) = (320,240), k_1 = \mathbf{0.5}, k_2 = \mathbf{1}$. Wizard images achieve superior performance over random images on all intrinsic parameters. Considering the autocorrelation matrices can further provide the most accurate and precise estimation outcomes.}
\label{fig:syn-test2}
\end{figure}

However, in this experiment, we notice that our system does not show much advantage over the randomly-taken images of the estimated distortion coefficients $k_1$ and $k_2$.  
Thus, a second experiment is performed with larger radial distortion coefficients $k_1 = 0.5$ and $k_2 = 1$, while the focal length and principal point stay the same. 
Fig.~\ref{fig:syn-test2} shows the effectiveness of the proposed system, especially with the consideration of autocorrelation matrix for target points.
When the radial distortion is large, we notice that not only both distortion coefficients, but also the focal length and principle points (not shown here) estimated from purely random images deviate much from the ground truth, as was also reported in~\cite{weng1992camera}. 
In contrast, our system still manifests the ability of centering around the ground truth with incomparably low standard deviation.  
Furthermore, compared to the simple case of the proposed system, both first-order statistics features appear to be most desirable when considering the autocorrelation matrices for the feature points.
%using only random images to estimate the focal length has fairly inferior performance 

\begin{figure}[!ht]
  \centering
  \begin{tabular}{cc}
    \includegraphics[width = 0.48\linewidth]{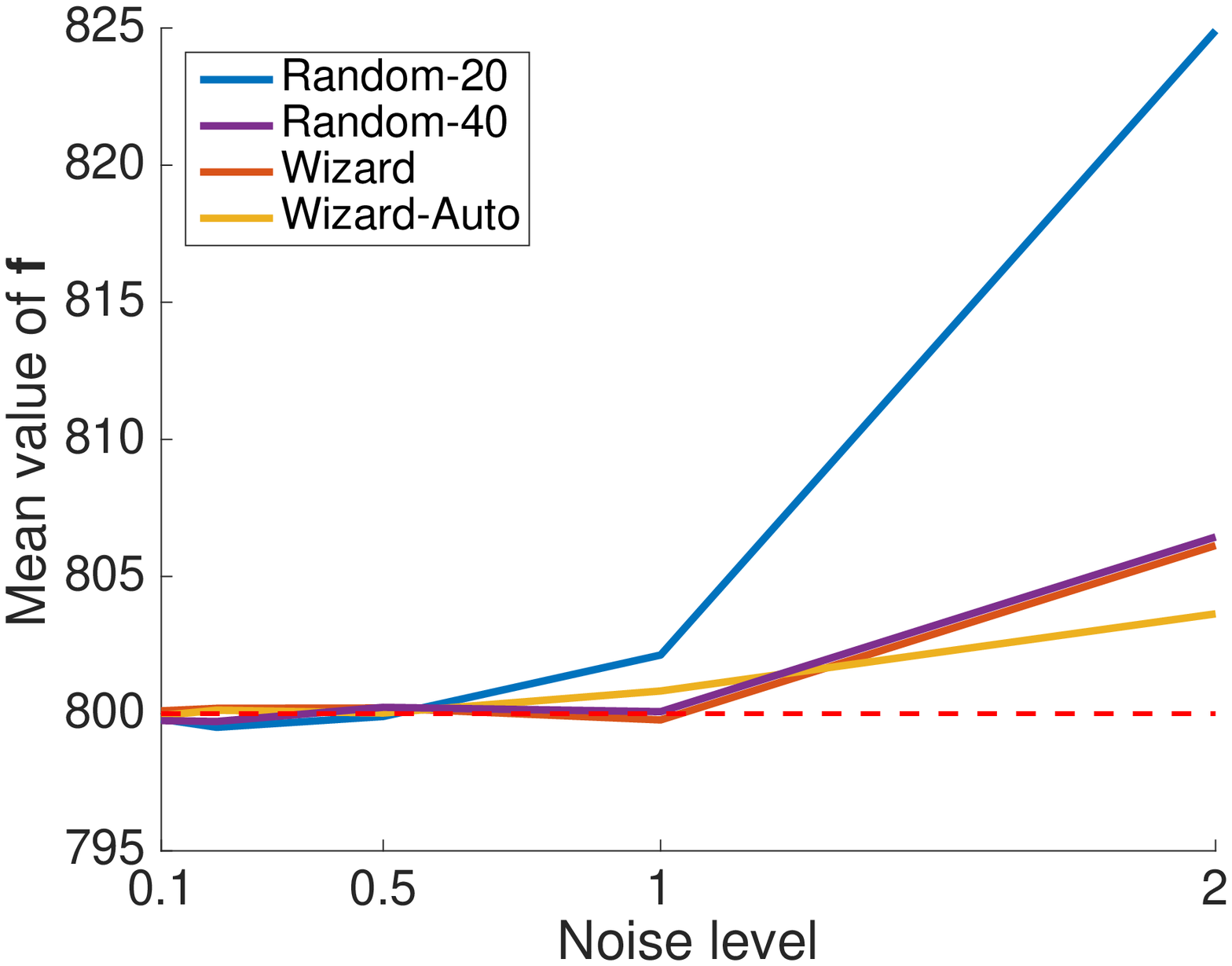}  &
    \includegraphics[width = 0.48\linewidth]{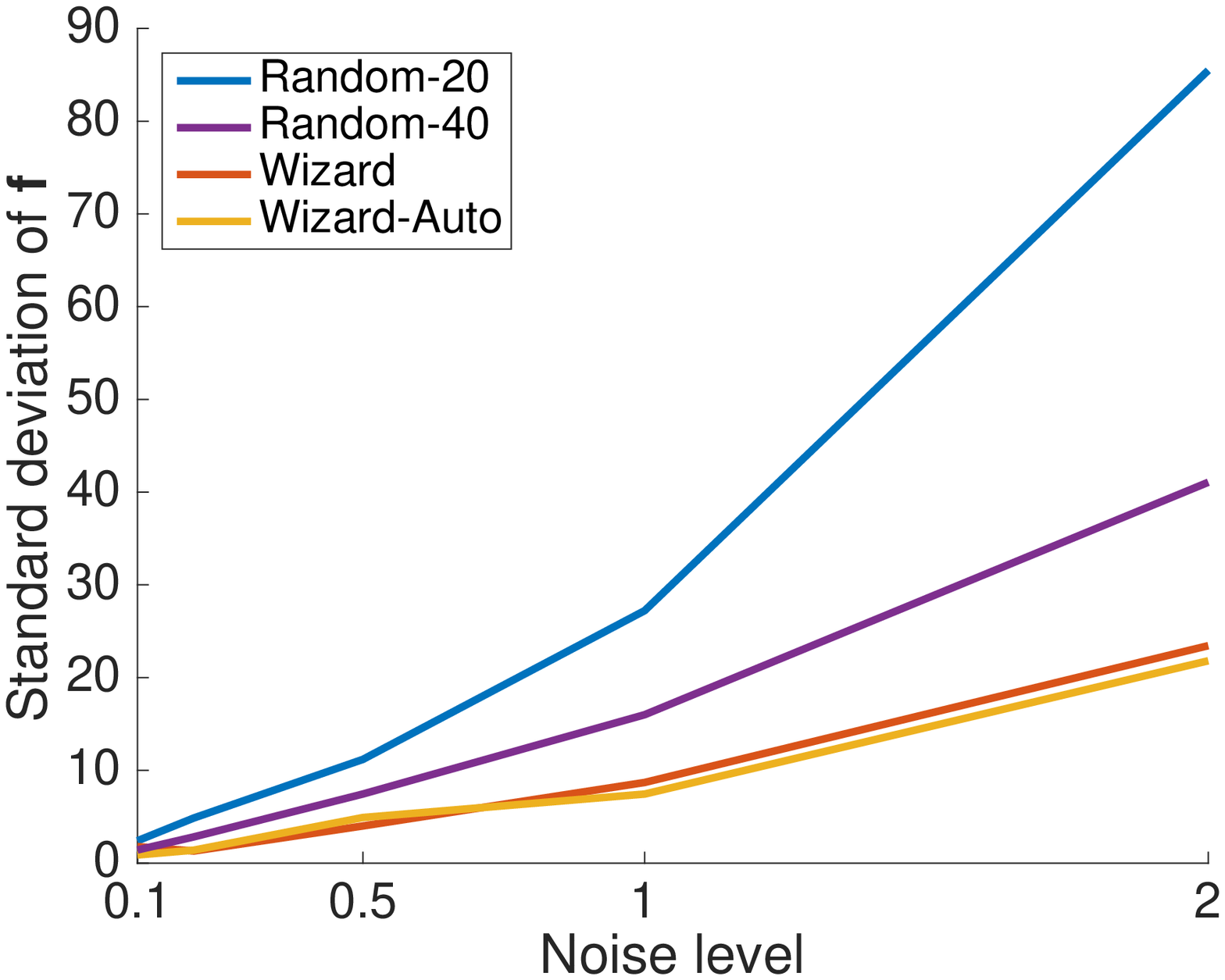}
  \end{tabular}
\caption{Comparisons among various calibration schemes of the robustness to noise. Zero-mean Gaussian noise with standard deviation of $0.1, 0.2, 0.5, 1$ respectively $2$ pixels is added to 2D target points. The focal length estimated from both of our methods with only 20 images, is more accurate (left) and precise (right) than that from 40 random images, especially when the noise level is high.}
\label{fig:syn-test3}
\end{figure}

\textbf{Robustness to noise.}
We are also interested in the performance of our approach with respect to the level of noise added to 2D corner points.
In this experiment, we compare 4 different configurations: $20$ random images, $40$ random images, $3$ random + 17 wizard images and $3$ random + 17 wizard-Auto images.
Zero-mean Gaussian noise with standard deviation of $0.1, 0.2, 0.5, 1$ with respect to $2$ pixels has been added to the image points respectively, and the comparisons are shown in Fig.~\ref{fig:syn-test3}.
Specifically, it can be distinctly seen from the figure that, even when unrealistically strong noise is added ($\sigma = 2$), both versions of our approach (3 random + 17 wizard images) still provide better accuracy than even 40 random images.
More synthetic experiments can be found in the supplementary material.

%%%%%%%%%%%%%%%%%%%%%%%%%%%%%%%%%%%%%%%%%%%%%%%%%%%

\subsection{Real-world evaluations}
\label{sec:result-real}

\begin{figure*}[!ht]
\centering
\includegraphics[width = 0.18\linewidth]{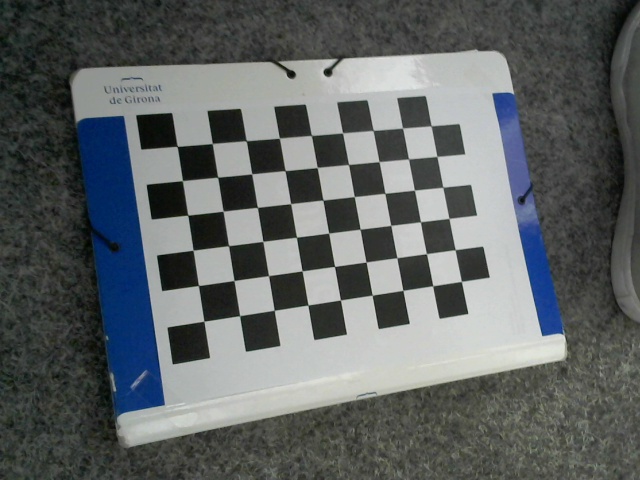}
\includegraphics[width = 0.18\linewidth]{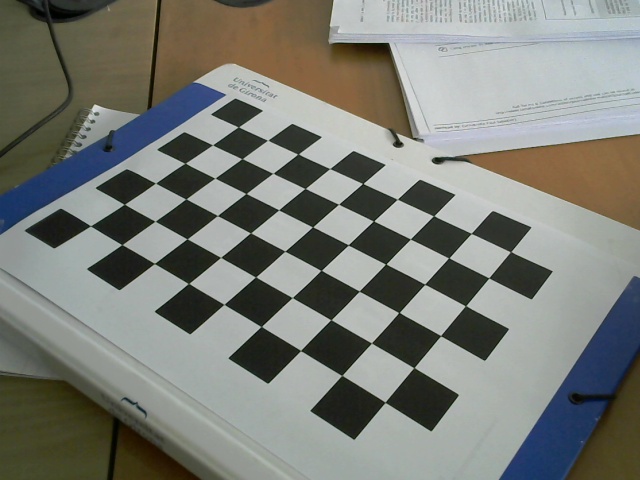}
\includegraphics[width = 0.18\linewidth]{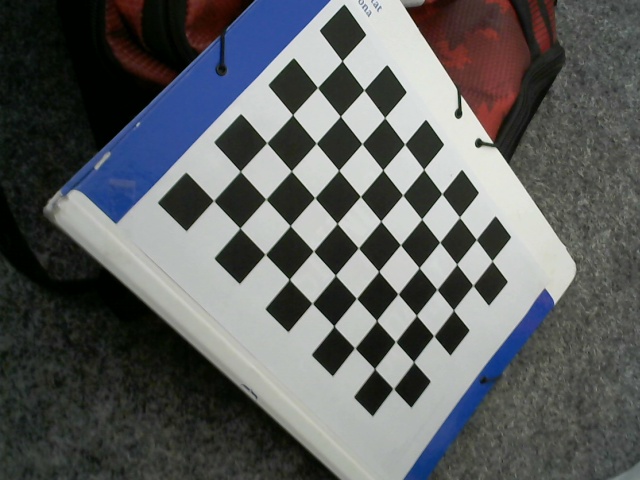}
\includegraphics[width = 0.18\linewidth]{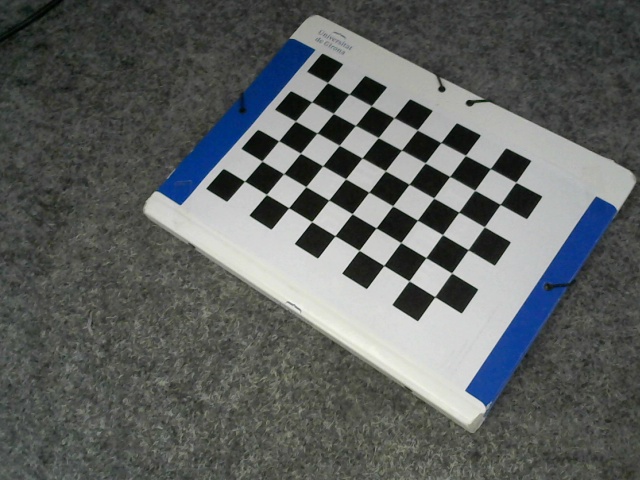}
\includegraphics[width = 0.18\linewidth]{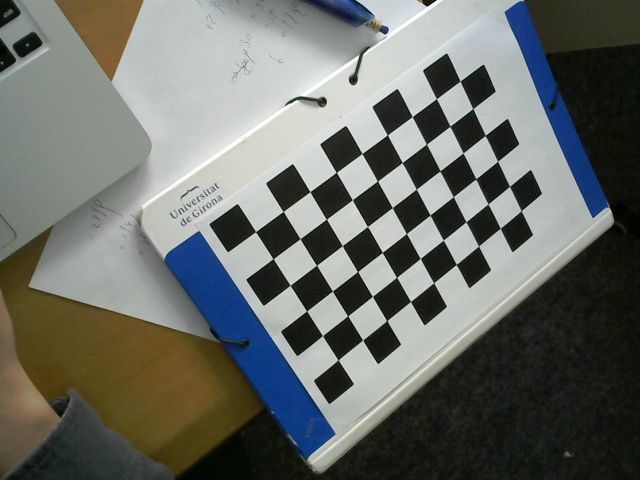}\\
\includegraphics[width = 0.18\linewidth]{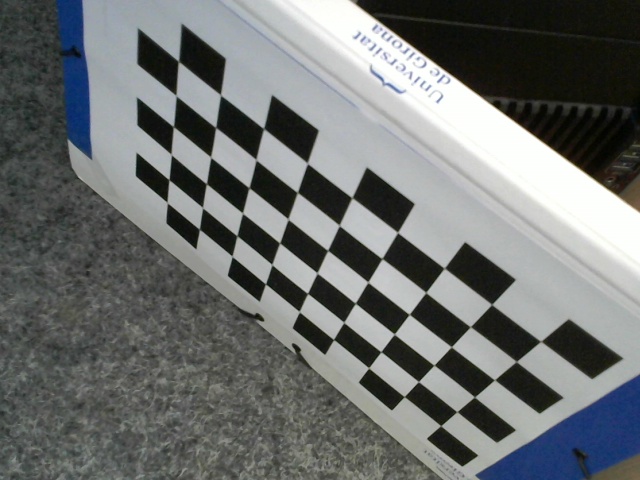}
\includegraphics[width = 0.18\linewidth]{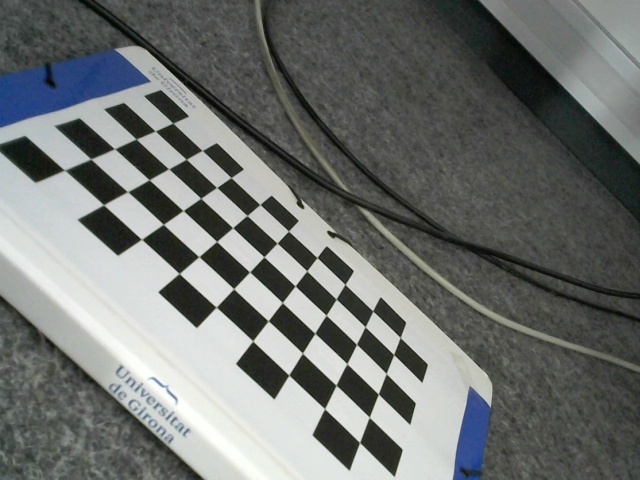}
\includegraphics[width = 0.18\linewidth]{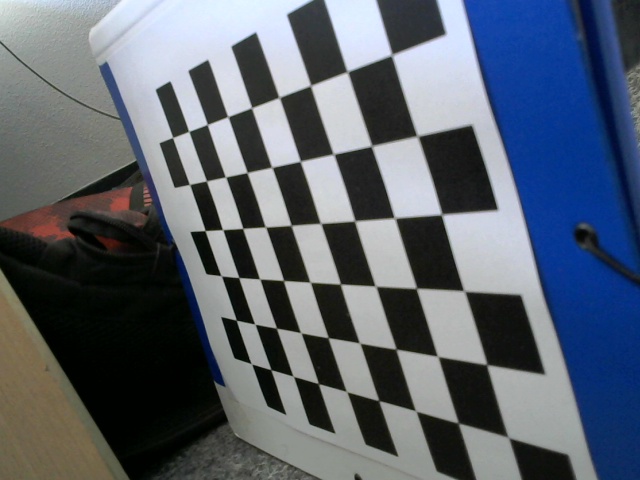}
\includegraphics[width = 0.18\linewidth]{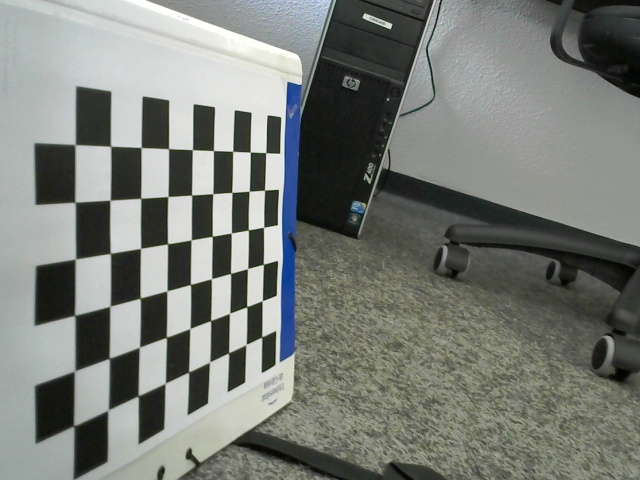}
\includegraphics[width = 0.18\linewidth]{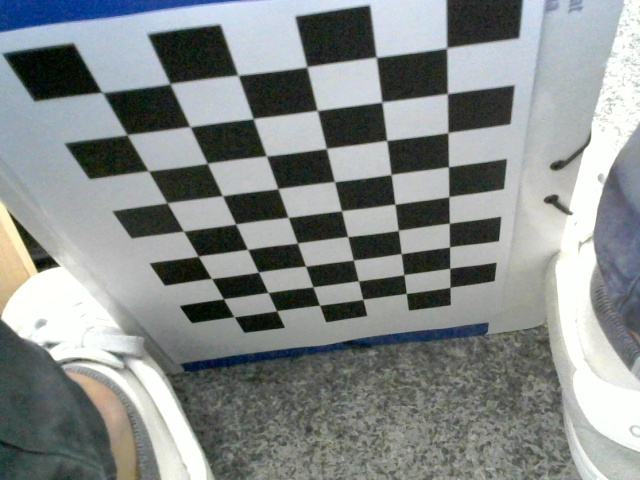}
\caption{Sample images used for the calibration in real-world tests. Top row: freely-taken images. Bottom row: wizard guided images.}
\label{fig:img-example}
\end{figure*}

Although the performance of the Calibration Wizard has been demonstrated for the synthetic data, we ultimately want to evaluate the effectiveness of its proposed next best pose on real-world examples.
We designed two experiments for this purpose where we also compare with calibrations obtained with freely taken images. Evaluating calibration results is difficult since ground truth is not readily available; we thus devised two experiments where calibration quality is assessed through evaluating results of applications -- pose estimation and SfM.
We used the commonly-used Logitech C270H HD Webcam in our experiments. It has an image size of $640\times 480$ and around $60^\circ$ field of view.
Fig.~\ref{fig:img-example} provides some sample calibration images.
One may notice that wizard-suggested images indeed correspond to poses often chosen by experts for stable calibration: large inclination of the target along the viewing direction, targets covering well the field of view and/or reaching the image border.

In the following, we denote ``$x$-free'' the calibration results from $x$ images acquired freely by an experienced user using OpenCV, compared to ``$x$-wizard'' where guidance was used. 
% Note that we could not compare our method with the similar method AprilCal~\cite{richardson2013aprilcal} because it seems out of maintenance and we were not able to compile and run it. 

% I think we should not put the following sentence. While it is indeed more practical for people to have something on top of openCV, it's not a reason not to compare. The other reason is, AprilCal could not be easily adapted to OpenCV, while our method is built right upon OpenCV. To be fair, it is reasonable to compare only with the OpenCV method.

\begin{figure}[t]
  \centering
  \begin{tabular}{cc}
    \includegraphics[width = 0.44\linewidth]{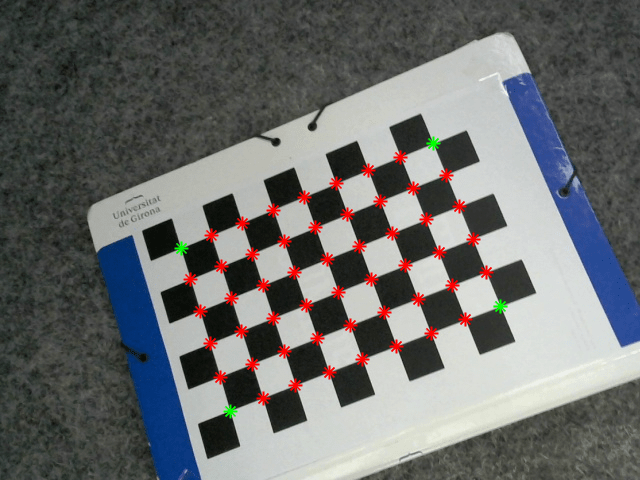} &
    \includegraphics[width = 0.47\linewidth,frame]{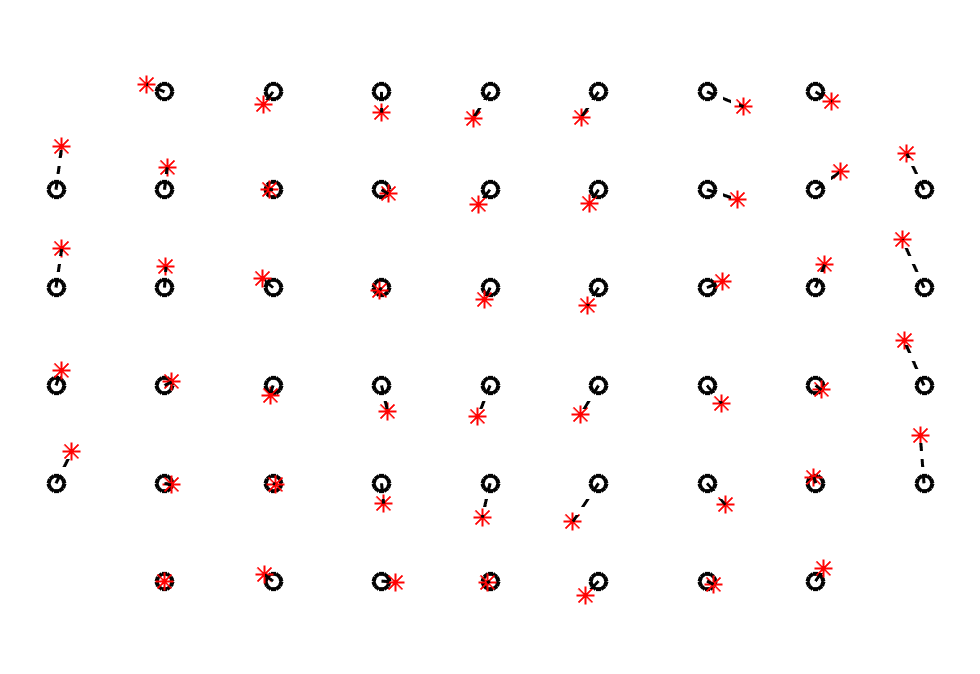}
  \end{tabular}
\caption{Pose estimation test. Left: checkerboard image where $4$ {\color{green}green} corner points are used for pose estimation, and the remaining $50$ {\color{red}red} points for reconstruction. Right: $50$ ground-truth points in black and residuals between them and the reconstructed corner points in {\color{red}red} (enlarged 50 times for visualization).}
\label{fig:pnp}
\end{figure}

\noindent\textbf{Pose estimation.}
Similar to the experiment performed in~\cite{ha2017deltille}, in order to quantitatively evaluate the quality of camera calibration, we design the first real-world experiment where,
apart from the images used for calibration, we then also acquire a number of extra checkerboard images which are only used for evaluation,
cf.\ Fig.~\ref{fig:pnp}. 

First, $4$ corner points are utilized to calculate the pose with EP$n$P~\cite{lepetit2009epnp}, given the intrinsic parameters provided by the calibration.
Then, since we have assumed the $Z$ components of the target points to be $0$ in the world coordinate system, it is straightforward to back-project the remaining $50$ points to 3D, onto the target plane, using the calibrated intrinsic parameters and the computed pose (cf.\ Fig.~\ref{fig:pnp} right).
The smaller Euclidean distance between the back-projected and theoretical 3D points, the better the calibration.

There are 80 images in total for testing so we have $50\times 80 = 4,000$ points for assessment.
The mean and standard deviation of the 4,000 distance errors are applied as metric.
%Note that we also consider the median values for the sake of eliminating the effect of some mis-detected outliers as well as the uncertainty in EP$n$P.
Table~\ref{tab:3d-test} demonstrates that our system, when using only 15 images for calibration, still exceeds the performance of using 50 freely acquired images and exceeds that of using 20 such images by about 5\%.
This seemingly small improvement may be considered as significant since it may be expected that differences are not large in this experiment. Even with a moderately incorrect calibration, pose estimation from 4 outermost target points will somewhat balance the reconstruction errors for the inner corner points.

\begin{table}[!htb]\centering
\caption{Pose estimation test with Logitech C270H HD Webcam.}
\footnotesize
\begin{tabular}{c|c|c||c|c|c}
\hline
        & mean  & std   &                           & mean           & std \\ \hline \hline
\; 3-free  & 0.856 & 1.130 & 3-free + 4-wizard           & 0.862          & 1.155          \\
10-free & 0.815 & 1.115 & 3-free + 6-wizard           & 0.783          & 1.092          \\
20-free & 0.802 & 1.115 & 3-free + 9-wizard           & 0.788          & 1.104          \\
50-free & 0.789 & 1.108 & \textbf{3-free + 12-wizard} & \textbf{0.763} & \textbf{1.082} \\ \hline
\end{tabular}
\label{tab:3d-test}
% PS-XXX: I deleted the median values, the mean values are actually more consistent I think.
\end{table} 

We also tested our approach on the FaceTime HD camera of a MacBook Pro. This camera has higher resolution and different field of view compared to other commonly use webcams, so it is a suitable alternative to show the robustness of our method. As shown in Table~\ref{tab:3d-test-mac}, adding only one or two wizard images can largely reduce the Euclidean distance and outperforms the results from freely taking many more images.

\begin{table}[!htb]\centering
\caption{Pose estimation test with FaceTime HD camera.}
\footnotesize
\begin{tabular}{c|c|c||c|c|c}
\hline
        & mean  & std   &                           & mean           & std \\ \hline \hline
\; 3-free  & 2.503 & 2.557 & 3-free + 1-wizard      & 1.455          & 1.630          \\
10-free & 1.664 & 1.839 & \textbf{3-free + 2-wizard}           & \textbf{1.165}          & \textbf{1.491}          \\
20-free & 1.255 & 1.606 & & &          \\\hline
\end{tabular}
\label{tab:3d-test-mac}
\end{table}

% \begin{figure}[!ht]
% \centering
% \includegraphics[width = 0.32\linewidth]{img/rand1.jpg}
% \includegraphics[width = 0.32\linewidth]{img/rand3.jpg}
% \includegraphics[width = 0.32\linewidth]{img/rand4.jpg}\\
% \includegraphics[width = 0.32\linewidth]{img/wizard1.jpg}
% \includegraphics[width = 0.32\linewidth]{img/wizard2.jpg}
% % \includegraphics[width = 0.2\linewidth]{img/wizard3.jpg}
% \includegraphics[width = 0.32\linewidth]{img/wizard5.jpg}
% \caption{Sample images used for the calibration in real-world tests. Top row: freely-taken images. Bottom row: wizard guided images.}
% \label{fig:img-example}
% \end{figure}

\begin{figure}[!htb]
\centering
    \includegraphics[width = 0.9\linewidth]{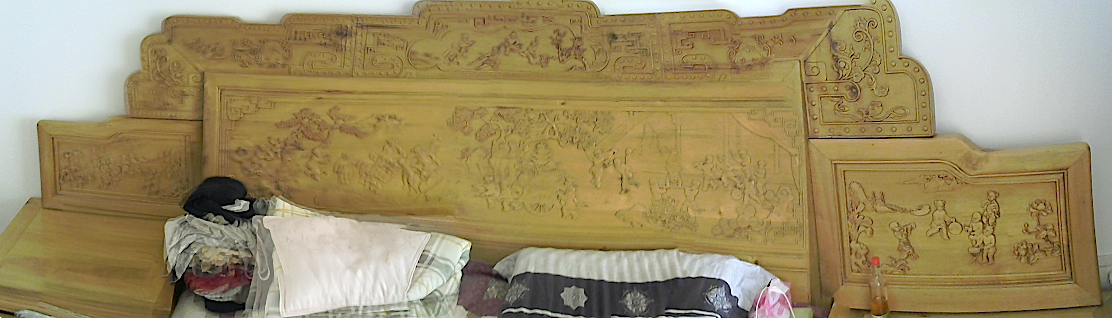} 
    \includegraphics[width = 0.9\linewidth]{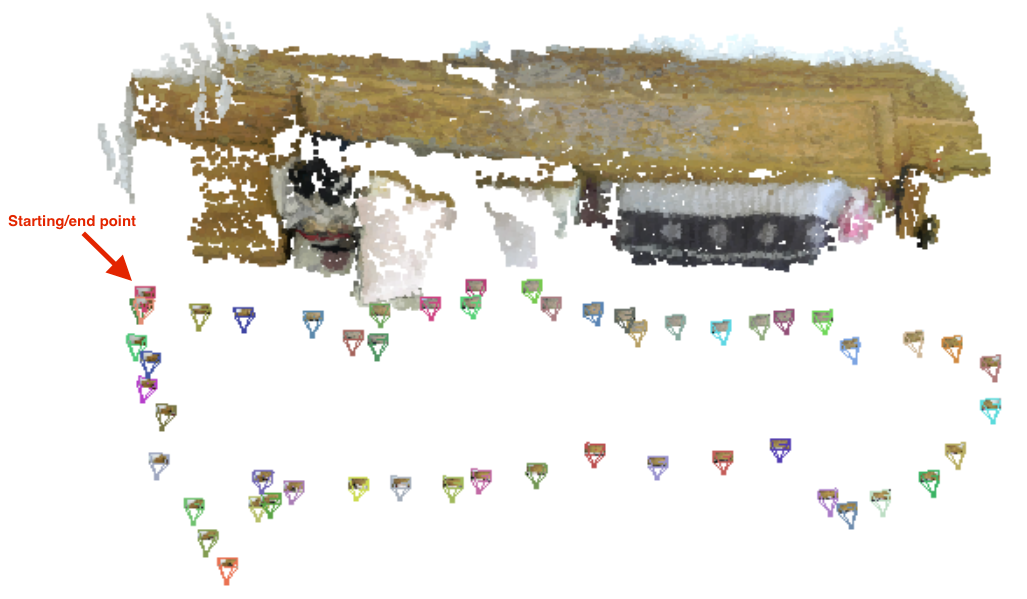} 
\caption{Structure from motion test.
Top: Panorama stitched by Hugin (\url{http://hugin.sourceforge.net}), showing the test scene.
Bottom: Result of applying VisualSfM~\cite{wu2013towards} to build a 3D model.
We started capturing images from the left and moved clockwise, finally came back approximately to the starting point.}
\label{fig:sfm}
\end{figure}

\noindent\textbf{Structure from Motion test.}
In this last experiment, we assess our Calibration Wizard by investigating the quality of 3D reconstruction in a structure-from-motion (SfM) setting. 
The object to be reconstructed is the backrest of a carved wooden bed, as shown in Fig.~\ref{fig:sfm}.
We devised a simple but meaningful experiment to evaluate the quality of the calibration, as follows.
We captured images from the far left of the object and gradually move to the right side, and then proceed backwards and return to the left, approximately to the starting point.
The acquired images are then provided as input to VisualSfM~\cite{wu2013towards}; we added an identical copy of the first image of the sequence, to the end of the sequence, but without ``telling'' this to the SfM tool and without using a loop detection method during SfM.
The purpose of doing so is: if calibration is accurate, the incremental SfM should return poses for the first image and the added identical last image, which are close to one another.
Measuring the difference in pose is not sufficient since the global scale of the reconstruction can be arbitrarily chosen by the SfM tool for each trial.
So instead, we project all 3D points that were reconstructed on the basis of interest points extracted in the first image, using the pose computed by SfM for the identical last image, and measure the distance between the two sets of 2D points such constructed.
This distance is independent of the scene scale and is thus a good indicator of the quality of the SfM result which in turn is a good indicator of the quality of the calibration used for SfM.

% VisualSfM~\cite{wu2013towards} has been used for the baseline method to reconstruct the 3D model illustrated in Fig.~\ref{fig:sfm}. 
Note that we only match two consecutive frames instead of full-pairwise matching within the given sequence.
In this case, 2D errors are accumulated so the reconstruction results highlight the calibration accuracy more strongly.

The experiment is described as follows.
We first obtain the 5-parameter calibration result (including two radial distortion coefficients), from 3 freely acquired images (``3-free'').
Then, on the one hand, another 17 images are taken, from which the intrinsic parameters of ``7-free'' and ``20-free'' are obtained.
On the other hand, we take another 4 sequential images proposed by the Calibration Wizard, where we get the intrinsic parameters of ``3-free + 2-wizard'' and ``3-free + 4-wizard''.
And now, we load VisualSfM with intrinsic parameters of these five configurations respectively, along with the backrest sequence taken by the same camera.
It is worth mentioning that we conduct five trials of VisualSfM for each configuration in order to lessen the influence of the stochastic nature of the SfM algorithm.

% \begin{table}[!ht]
% \centering
% \caption{2D errors of SfM tests under different calibration schemes.}
% \begin{tabular}{c|c|c|c|c}
% \hline
%               & mean            & std            & median & uncertainty range        \\\hline \hline
% $3$ rand         & 43.575          & 11.547         & 44.318       & 100-500   \\  
% $7$ rand   &30.527 & 11.682 & 31.728 & 50-330 \\
% $20$ rand        & 15.712          & 10.512         & 16.131      & 20-140    \\
% $3$ rand $+$ $2$ wizard \quad & 17.391          & 10.757         & 13.158    & 20-45      \\
% $3$ rand $+$ $4$ wizard \quad & \textbf{14.448} & \textbf{9.056} & \textbf{10.564}  & \textbf{14-34} \\\hline      
% \end{tabular}
% \label{tab:sfm}
% \end{table}
%
\begin{table}[t]
\centering
\caption{2D errors of SfM tests under various calibration schemes.}
\begin{tabular}{c||r@{}l|r@{}l|r@{}l}
\hline
Calibration scheme & \multicolumn{2}{c|}{mean} & \multicolumn{2}{c|}{std} & \multicolumn{2}{c}{median} \\
\hline \hline
3-free                      & 43 & .6 & 11 & .5 & 44 & .3 \\  
7-free                      & 30 & .5 & 11 & .7 & 31 & .7 \\
20-free                     & 15 & .7 & 10 & .5 & 16 & .1      \\
3-free $+$ 2-wizard \quad & 17 & .4 & 10 & .8 & 13 & .2    \\
3-free $+$ 4-wizard \quad & \textbf{14} & \textbf{.4} & \textbf{9} & \textbf{.1} & \textbf{10} & \textbf{.6}  \\
\hline      
\end{tabular}
\label{tab:sfm}
\end{table}
Results are listed in Table~\ref{tab:sfm}, where we evaluate the 2D errors across all 5 trials.
% Besides, we also include the range of uncertainty maps acquired from every case. 
Some observations can be made:
with only 5 images in total, ``3-free + 2-wizard'' has already provided an accuracy competitive to 20 freely-taken images.
Both ``7-free'' and ``3-free + 4-wizard'' use 7 images for calibration, but it can be clearly noticed that the latter one has far lower errors in all aspects.
% \begin{itemize}
% 	\item With only 5 images in total, ``3 rand + 2 wizard'' has already provided competitive accuracy as 20 randomly-taken images.
% 	\item Both ``7 rand'' and ``3 rand + 4 wizard'' use 7 images for calibration, but it can be clearly noticed that the latter one has far lower errors in all aspects.
% 	\item As for the range of uncertainty maps, we can aware that our system indeed has the capability of reducing uncertainty significantly.
% \end{itemize}
It is reasonable to conclude that our method notably improves the quality of calibration and 3D reconstruction with a considerably small number of calibration images.

Finally, it is worth mentioning that all real-world experiments were performed with a 2.7 GHz Intel i5 CPU (no GPU used). To compute the next best pose with a $9\times6$ target, our un-optimized C++ code took about $0.4s$ for 3 target images and $1.5s$ for 15 images (increasing roughly linearly per image), but we found that 10 images are usually sufficient for a good calibration. 

%--------------------------------------------------------------------------
\section{Conclusions}
\label{sec:conclusions}

Calibration Wizard is a novel approach which can guide any user through the calibration process.
We have shown that accurate intrinsic parameters can be obtained from only a small number of images suggested by this out-of-the-box system.
Some ideas for future work were already mentioned in section \ref{sec:autocorrelation.next}.
We also plan to apply the approach to very wide field of view cameras such as fisheyes.

{\small
\bibliographystyle{ieee_fullname}
\bibliography{egbib}
}

\clearpage

%\pagenumbering{gobble}
\setcounter{section}{0}
\setcounter{figure}{0}
\setcounter{table}{0}
\setcounter{equation}{0}

\include{supplementary}

\end{document}

%% file: supplementary.tex
\title{\vspace{-1em}Calibration Wizard: A Guidance System for Camera Calibration\\Based on Modelling Geometric and Corner Uncertainty\\-- Supplementary Material --}

\author{Songyou Peng$^*$\\
ETH Zurich\\
{\tt\small songyou.peng@inf.ethz.ch}
\and
Peter Sturm\\
INRIA Grenoble -- Rh\^{o}ne-Alpes \\
{\tt\small peter.sturm@inria.fr}
}

\maketitle
In this document, we provide the following details:
\begin{itemize}
\item Section \ref{s.experiments}: provides another synthetic experiment to further demonstrate the effectiveness of our system.
\item Section \ref{s.computations}: efficient computation of the next best pose (cf. section 2.2 of the main paper) including the incorporation of autocorrelation matrices in this computation (section 3 of the main paper).
\item Section \ref{s.derivatives}: full details on the computation of partial derivatives in $J$ (cf. sections 2.1 and 2.2 of the main paper) and a full example of this: pinhole model with radial distortion.
\item Section \ref{sec:uncertainty-map}: describes a way of visualizing the uncertainty of calibration results through pixel-wise uncertainty maps, displaying expected reprojection errors induced by errors on intrinsic parameters.
\end{itemize}

\section{Another Synthetic Experiment}
\label{s.experiments}
When users try to calibrate a camera, they tend to randomly move the camera around and get as many images as possible for calibration. With our calibration wizard, we have already shown the superiority over such randomly-captured case in the main paper.
Now, considering that calibration is an interactive procedure, i.e., a user can move the camera around, we could actually keep all the intermediate frames when moving from one position to the next, and then all these frames can be used for calibration. 
Here we perform another synthetic test to validate that the calibration can still be improved by moving to optimal poses under this scenario.
Provided the frame rate is $25$ fps and it takes 1 second to move from one pose to the next, 25 images can then be acquired within \textit{one path}.
To this end, the experimental process is as follows.

First, we create 4 random poses using exactly the same way described in the main paper, from which we could acquire 3 paths by pose interpolation. Thus, there are $3\times 25 = 75$ frames in total for the case ``Random''. This process mimics moving the camera from one pose to another continuously for three times.

Then, we randomly take 5 images within the first path to get the initial calibration for our wizard. The wizard then proposes a next best pose and while the simulated camera moves there, 25 additional images for calibration are acquired. This is done 3 times, for 75 additional calibration images in total.
This input leads to the results labeled as ``Wizard'' in the following.
The same process is also applied for taking autocorrelation matrix into account in next best pose computation (``Wizard-Auto'')

\begin{figure}[!ht]
  \centering
  \begin{tabular}{cc}
    \includegraphics[width = 0.48\linewidth]{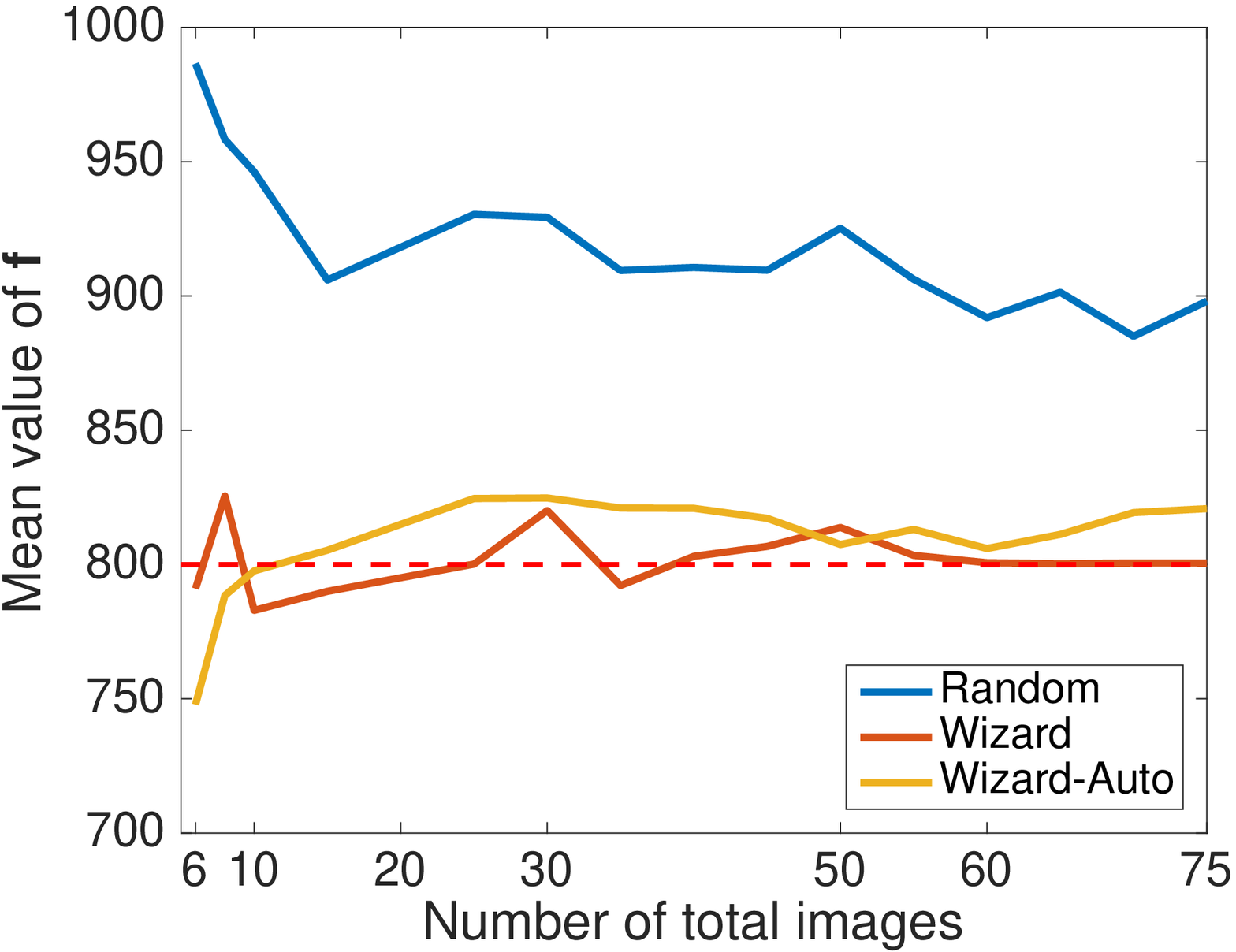}  &
    \includegraphics[width = 0.48\linewidth]{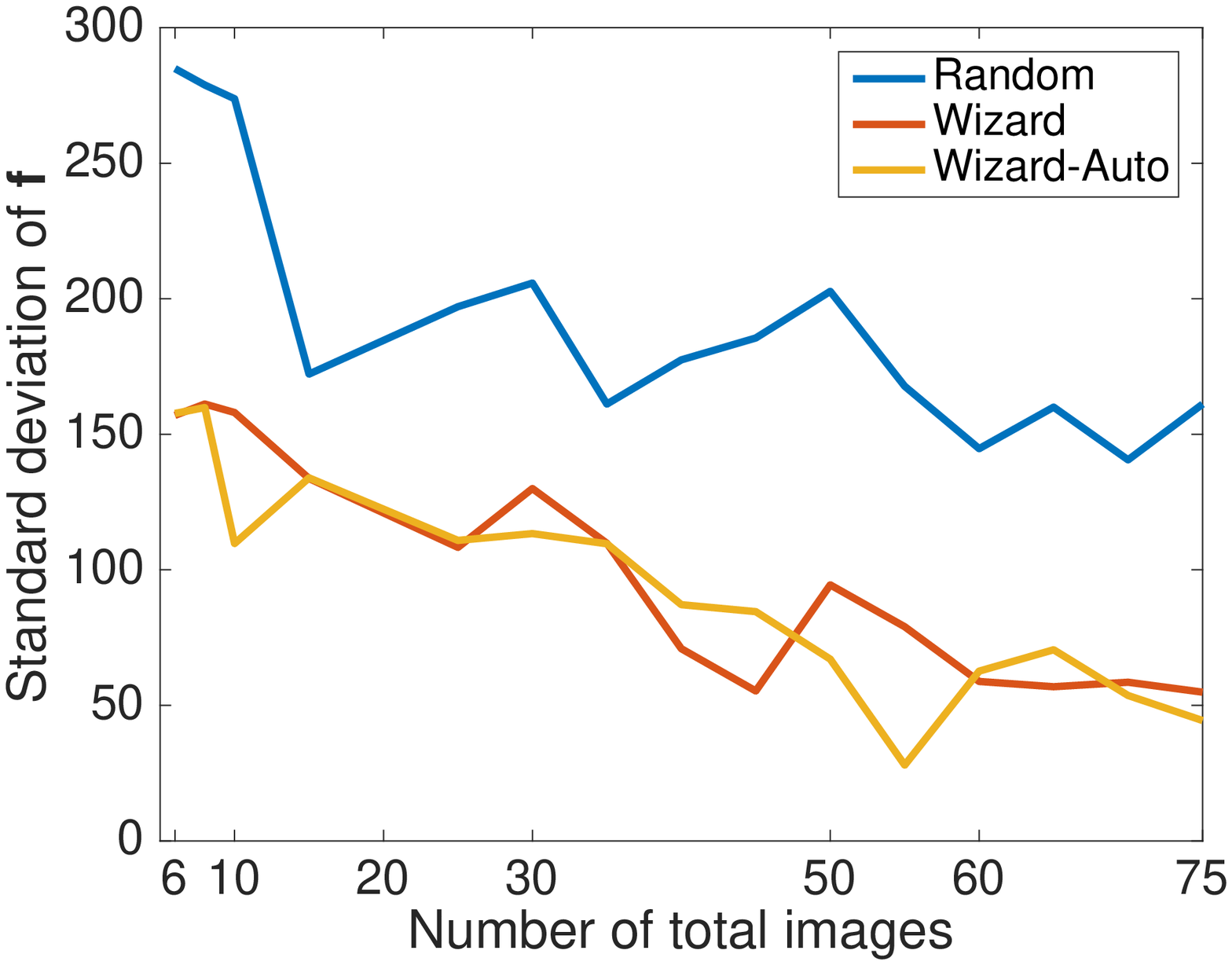}  \\         
    \includegraphics[width = 0.48\linewidth]{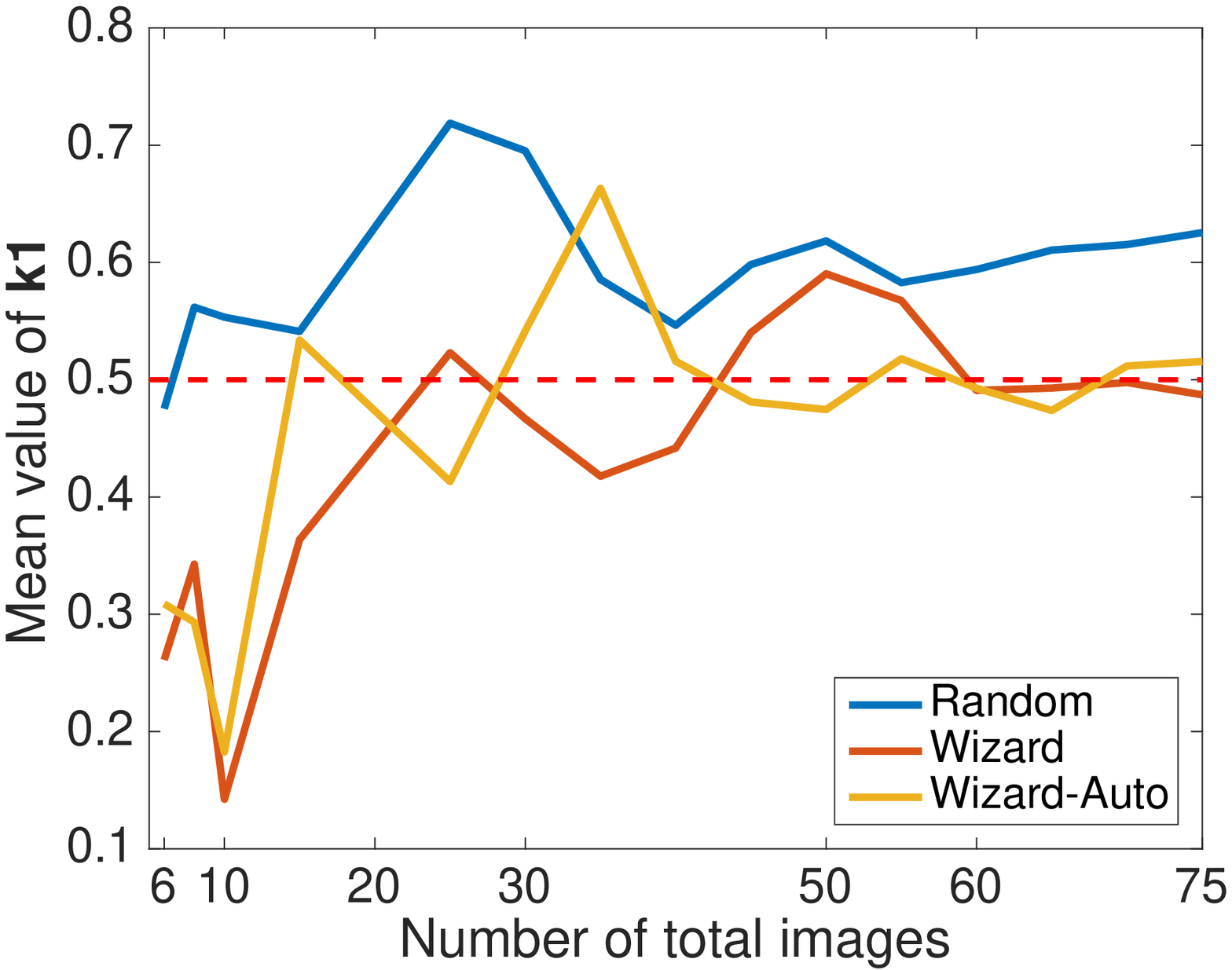}  &
    \includegraphics[width = 0.48\linewidth]{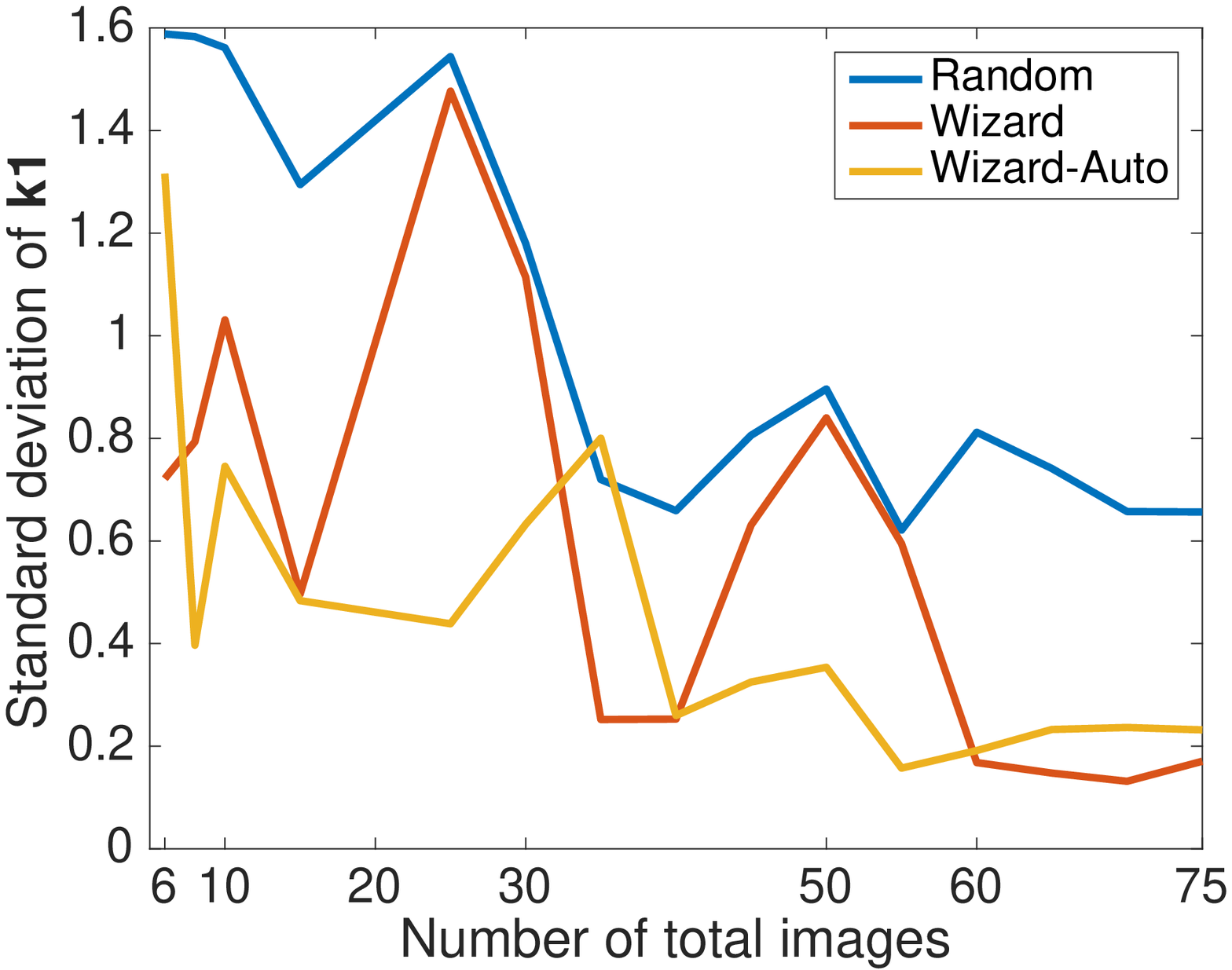}  \\             
    \includegraphics[width = 0.48\linewidth]{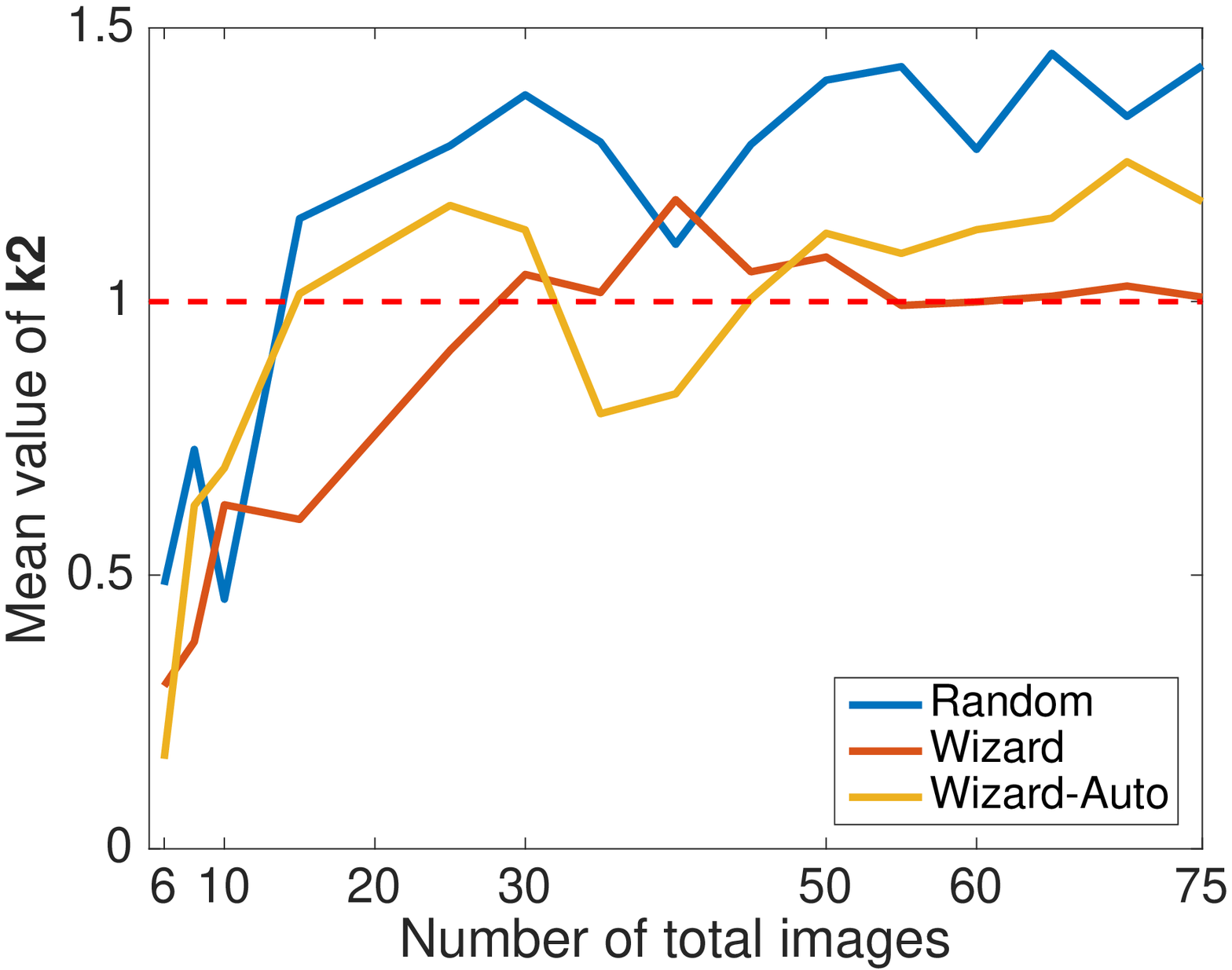}  &
    \includegraphics[width = 0.48\linewidth]{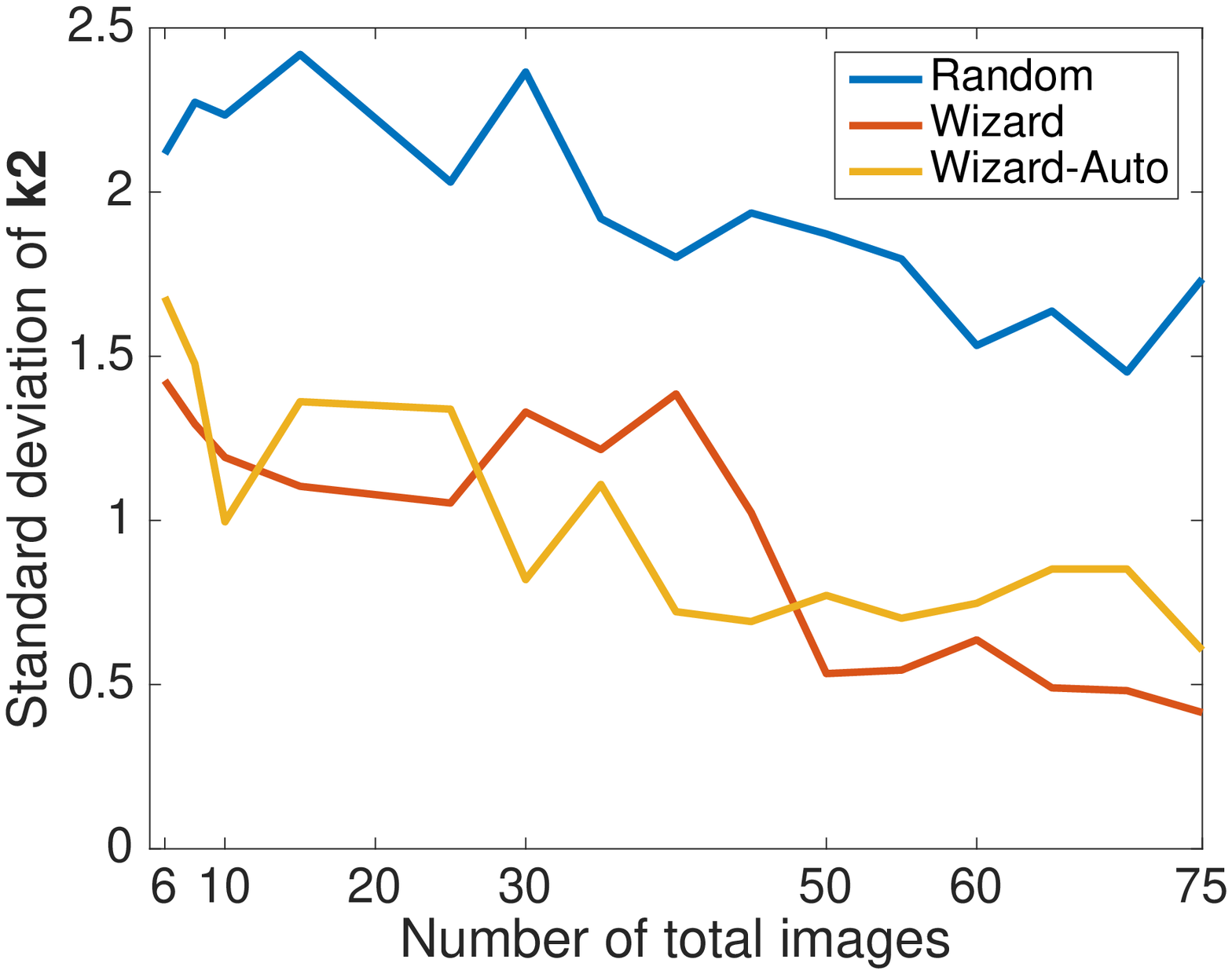}  \\                 
  \end{tabular}
\caption{Comparisons of the intrinsic parameters estimated from three schemes on synthetic data: the paths generated from random poses, generated from the poses proposed by calibration wizard without and with autocorrelation matrix. $f = 800, (u,v) = (320,240), k_1 = \mathbf{0.5}, k_2 = \mathbf{1}$. {\color{red} Red} dashed lines represent the ground truth values. Wizard images achieve superior performance over random images on all intrinsic parameters.}
\label{fig:syn-test21_supp}
\end{figure}

Finally, we simply compare the calibration results acquired from random-generated paths and wizard-generated ones. 
As illustrated in Fig.~\ref{fig:syn-test21_supp}, we can clearly notice that the calibration results from our system have higher precision and accuracy than the results from random paths.

\begin{figure}[!ht]
  \centering
  \begin{tabular}{cc}
    \includegraphics[width = 0.48\linewidth]{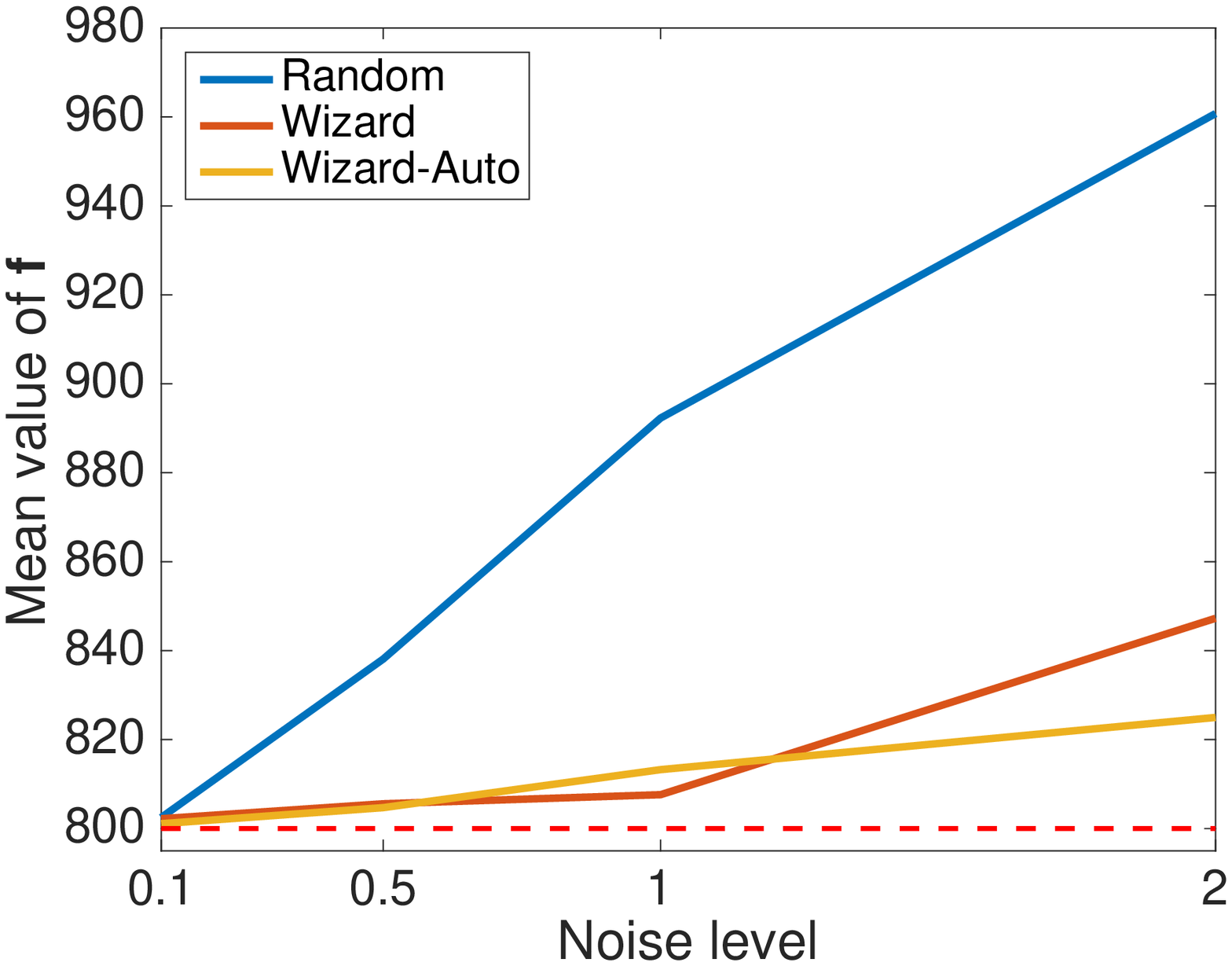}  &
    \includegraphics[width = 0.48\linewidth]{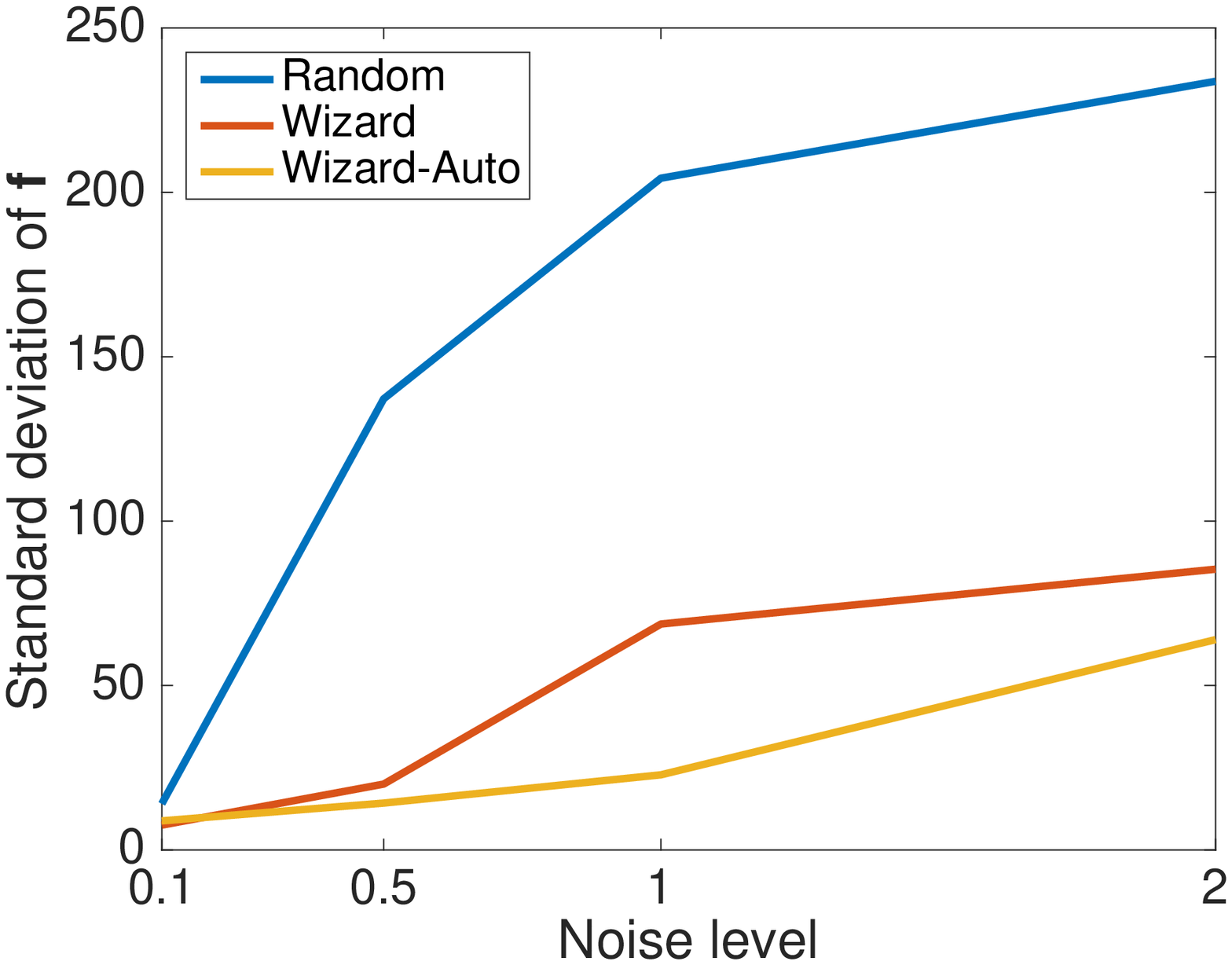}
  \end{tabular}
\caption{Comparisons of the calibration schemes concerning the robustness to added noise. Zero-mean Gaussian noise with standard deviation of $0.1, 0.2, 0.5, 1$ respectively $2$ pixels is added to the 2D target points. The intrinsic parameters estimated from the wizard-generated paths (shown are results for the focal length), are more accurate (left) and precise (right) than random-generated paths, especially when the noise level is high. The total number of image is 75.}
\label{fig:syn-test3_supp}
\end{figure}

We also perform the comparison with respect to the level of noise added to 2D corner points.
In this experiment, zero-mean Gaussian noise with standard deviation of $0.1, 0.2, 0.5, 1$ respectively $2$ pixels has been added to corner points, and the comparisons are shown in Fig.~\ref{fig:syn-test3_supp}.
Again, even when unrealistically strong noise is added ($\sigma = 2$), our system provides much better estimation of the focal length. Moreover, when considering the autocorrelation matrices, our system seems more robust to the noise.

Note that we only select the results for some intrinsic parameters for these two experiments, but similar results can be expected for other parameters like principal points.

% \begin{figure}[t]
%   \centering
%   \begin{tabular}{cc}
%     \includegraphics[width = 0.48\linewidth]{img/synF_mean_small}  &
%     \includegraphics[width = 0.48\linewidth]{img/synF_std_small}
% %    \multicolumn{2}{c}{\includegraphics[width = 0.48\linewidth]{img/synCompZ}}\\
% %    \multicolumn{2}{c}{\small (c)}
%   \end{tabular}
% \caption{Comparisons of the focal length estimated from three schemes on synthetic data: the paths generated from random poses, generated from the poses proposed by calibration wizard without and with autocorrelation matrix. $f = 800, (u,v) = (320,240), k_1 = 0.01, k_2 = 0.1$. Initial calibration for wizard was done with $5$ arbitrary images in the first random-generated path. Left: Mean values of the estimated focal length, where the {\color{red} red} dashed line represents the ground truth $f=800$. Right: Standard deviations of the estimated focal length. Wizard images provide significantly more accurate and precise calibration results than random ones.}
% \label{fig:syn-test11}
% \end{figure}

\section{An efficient way of computing the covariance matrix of intrinsic parameters}
\label{s.computations}

As already explained in the paper, the Jacobian matrix $J$ can be denoted as:
\begin{equation}
J = \begin{pmatrix}
A_1 & B_1 & 0 & \cdots & 0 \\
A_2 & 0 & B_2 & \cdots & 0 \\
\vdots & \vdots & \vdots & \ddots & \vdots \\
A_m & 0 & 0 & \cdots & B_m
\end{pmatrix}
\end{equation}

Let us examine in detail the incorporation of the autocorrelation matrices of image corner points in the computation of the information matrix, as shown in Eq.~(9) in the paper:
\begin{equation}
H = J^\top \text{diag}(C_{11}, C_{12}, \cdots, C_{1n}, C_{21}, \cdots, C_{mn}) J
\label{eq.information_matrix}
\end{equation}

Let us decompose the matrices $A_i$ and $B_i$ appearing in the definition of the $J$, into matrices $A_{ij}$ and $B_{ij}$ containing the residual associated with a single image point each (point $j$ in image $i$):
\[
A_{ij} = \begin{pmatrix}
\frac{\partial \hat{x}_{ij}}{\Theta_1} & \cdots & \frac{\partial \hat{x}_{ij}}{\Theta_k} \\
\frac{\partial \hat{y}_{ij}}{\Theta_1} & \cdots & \frac{\partial \hat{y}_{ij}}{\Theta_k}
\end{pmatrix}
\enspace
B_{ij} = \begin{pmatrix}
\frac{\partial \hat{x}_{ij}}{\Pi_{i,1}} & \cdots & \frac{\partial \hat{x}_{ij}}{\Pi_{i,6}} \\
\frac{\partial \hat{y}_{ij}}{\Pi_{i,1}} & \cdots & \frac{\partial \hat{y}_{ij}}{\Pi_{i,6}}
\end{pmatrix}
\]

The $A_{ij}$ are of size $2\times k$ ($k$ is the number of intrinsic parameters) and the $B_{ij}$ of size $2\times 6$.
We then have:
\[
A_i = \begin{pmatrix} A_{i1} \\ \vdots \\ A_{in} \end{pmatrix}
\enspace
B_i = \begin{pmatrix}  B_{i1} \\ \vdots \\ B_{in}
\end{pmatrix}
\]

Now, the information matrix $H$ of Eq.~\eqref{eq.information_matrix} can be structured as:
\begin{equation}
H = \begin{pmatrix}
    U & W\\ W^\top & V
\end{pmatrix}
\label{eq:infoMat_new}
\end{equation}

where 
\begin{eqnarray}
U & = & \sum_i U_i \\
\text{with}\ U_i & = & \sum_j A_{ij}^\top C_{ij} A_{ij} \\
W & = & \begin{pmatrix} W_1 & \cdots & W_m \end{pmatrix} \\
\text{with}\ W_i & = & \sum_j A_{ij}^\top C_{ij} B_{ij} \\
V & = & \begin{pmatrix} V_1 & & & \\ & V_2 & & \\ & & \ddots & \\ & & & V_m \end{pmatrix} \\
\text{with}\ V_i & = & \sum_j B_{ij}^\top C_{ij} B_{ij}
\end{eqnarray}

$U$ is a symmetric matrix of size $k\times k$, $W$ is of size $k \times (6m)$ (remember that $m$ is the number of images) and $V$ is a block-diagonal matrix consisting of $m$ symmetric $6\times6$ matrices $V_i$.

As described in~\cite{hartley2003multiple}, the upper-left sub-matrix of $H^{-1}$ is given by 
\begin{equation}
\Sigma = (U-W V^{-1} W^\top)^{+}
\label{eq:sigma_big}
\end{equation}
which is a pseudo-inverse of a $k\times k$ symmetric matrix.
Let us expand this using the above definitions of $U, V$ and $W$.

Using the above definitions of $U, V$ and $W$, we can rewrite $\Sigma$ as:
\begin{equation}
\Sigma = \left\{ \sum_i \left( U_i - W_i V_i^{-1} W_i^\top \right) \right\}^{+}
\label{eq.covariance}
\end{equation}

The computation is efficient.
The $V_i$ are symmetric and of size $6\times6$, hence their inversion is a small problem.
Other than that, the computation only involves products and sums of small matrices and one final inversion of a $k \times k$ symmetric matrix.

Let us now consider the computation of the next best pose.
As explained in the paper, we use global optimization algorithms for this purpose.
They will need to compute $\Sigma$ and its trace multiple times, for multiple hypothetical next poses.
It is thus interesting to study how to cut computation times by performing adequate pre-computations.
This is very simple in our case.
Let image $m+1$ be the next image, for which to compute the optimal pose.
Images $1$ till $m$ are thus already acquired and we have computed intrinsic and pose parameters for them.
All we need to do is to compute once the following part of equation (\ref{eq.covariance}), for images $1$ to $m$:
\begin{equation}
A = \sum_{i=1}^m \left( U_i - W_i V_i^{-1} W_i^\top \right)
\end{equation}

Then, for each hypothetical pose for image $m+1$, we only need to compute
\begin{equation}
\Sigma = \left\{ A + U_{m+1} - W_{m+1} V_{m+1}^{-1} W_{m+1}^\top \right\}^{+}
\end{equation}
This computation is efficient: besides the computation of partial derivatives of the projection function for the hypothetical new pose (see section \ref{s.derivatives} for details) and the sum or multiplication of small matrices (of size at most $(max(k,6) \times max(k,6)$), it only involves the inversion of a symmetric matrix of size $6\times6$ and of a symmetric matrix of size $k\times k$.

Note that in this section we handled the incorporation of autocorrelation matrices.
If one wishes to work without them, it suffices to replace the $C_{ij}$ by identity matrices.

%%%%%%%%%%%%%%%%%%%%%%%%%%%%%%%%%%%%%%%%%%%%%%%%%%%%%%%%%%%%%%%%%
\section{Details on the computation of partial derivatives in $J$}
\label{s.derivatives}

In the following, we use the following shorthand notations for partial derivatives, which makes for easier reading.
If $A$ is a vector of size $a$ and $B$ a vector of size $b$ (with possibly, $a=1$ or $b=1$), then we write:
\begin{equation}
\frac{\partial A}{\partial B}
=
\begin{pmatrix}
\frac{\partial A_1}{\partial B_1} & \cdots & \frac{\partial A_1}{\partial B_b} \\
\vdots & \ddots & \vdots \\
\frac{\partial A_a}{\partial B_1} & \cdots & \frac{\partial A_a}{\partial B_b}
\end{pmatrix}
\end{equation}

Also, if $A$ is a matrix and $B$ a scalar, then $\frac{\partial A}{\partial B}$ is the matrix gathering the partial derivatives of the coefficients of $A$ relative to the scalar $B$, in the same order as they appear in $A$.

As we have seen in the paper, $3D$ to $2D$ projections can be written as:
\begin{align}
	p_x(\Theta, \Pi, Q) &= q_x(\Theta, RQ+t) = q_x(\Theta, S)\\
	p_y(\Theta, \Pi, Q) &= q_y(\Theta, RQ+t) = q_y(\Theta, S)
\end{align}

Thanks to the chain rule, the derivatives of residuals $\hat{x}$ (for their definition, see section 2 of the main paper) with respect to extrinsic parameters can be decomposed as:
\begin{eqnarray}
\frac{\partial \hat{x}}{\partial R} & = &
\frac{\partial p_x}{\partial R} =
\frac{\partial q_x}{\partial S}\frac{\partial S}{\partial R} \label{eq.pd1} \\
\frac{\partial \hat{x}}{\partial t} & = & \frac{\partial p_x}{\partial t} = \frac{\partial q_x}{\partial S}\frac{\partial S}{\partial t} \label{eq.pd2}
\end{eqnarray}
and likewise for partial derivatives of $\hat{y}$.
Here, we use the informal notation
\begin{equation}
\partial R = \partial \begin{pmatrix} \alpha \\ \beta \\ \gamma \end{pmatrix}
\end{equation}
where the three angles $\alpha, \beta, \gamma$ parameterize the rotation matrix $R$ as shown on page 2 of the main paper and as reproduced here:
\begin{equation}
R = R_z(\gamma) R_y(\beta) R_x(\alpha)
\end{equation}

$\partial q_x/\partial S$ and $\partial q_y/\partial S$ depend on the projection functions for the camera model used for calibration (see below for an example, the pinhole model with one radial distortion coefficient), while $\partial S/\partial R$ and $\partial S/\partial t$ are generic for all camera models.
Remember that
\begin{equation}
S = RQ+t
\label{eq.S}
\end{equation}
Thus:
% \begin{equation}
% {
% \frac{\partial S}{\partial R} =
%   \begin{pmatrix}
%   \frac{\partial S_1}{\partial \alpha} & \frac{\partial S_1}{\partial \beta} & \frac{\partial S_1}{\partial \gamma} \\
%   \frac{\partial S_2}{\partial \alpha} & \frac{\partial S_2}{\partial \beta} & \frac{\partial S_2}{\partial \gamma} \\
%   \frac{\partial S_3}{\partial \alpha} & \frac{\partial S_3}{\partial \beta} & \frac{\partial S_3}{\partial \gamma}
%   \end{pmatrix}
%   = R
%   \begin{pmatrix}
%   Q_1 & Q_3 & -Q_2 \\
%   -Q_3 & Q_2 & Q_1 \\
%   Q_2 & -Q_1 & Q_3
%   \end{pmatrix}
%   }
% \end{equation}
% \begin{equation}
% \tiny
%     \frac{\partial S}{\partial t} =
%   \begin{pmatrix}
%   \frac{\partial S_1}{\partial t_1} & \frac{\partial S_1}{\partial t_2} & \frac{\partial S_1}{\partial t_3} \\
%   \frac{\partial S_2}{\partial t_1} & \frac{\partial S_2}{\partial t_2} & \frac{\partial S_2}{\partial t_3} \\
%   \frac{\partial S_3}{\partial t_1} & \frac{\partial S_3}{\partial t_2} & \frac{\partial S_3}{\partial t_3}
%   \end{pmatrix}
%   = I_{3\times3}   
% \end{equation}
\makeatletter
    \def\tagform@#1{\maketag@@@{\normalsize(#1)\@@italiccorr}}
\makeatother
{\small
\begin{align}
\label{eq:dSdR}
\frac{\partial S}{\partial R} &=
  \begin{pmatrix}
  \frac{\partial S_1}{\partial \alpha} & \frac{\partial S_1}{\partial \beta} & \frac{\partial S_1}{\partial \gamma} \\
  \frac{\partial S_2}{\partial \alpha} & \frac{\partial S_2}{\partial \beta} & \frac{\partial S_2}{\partial \gamma} \\
  \frac{\partial S_3}{\partial \alpha} & \frac{\partial S_3}{\partial \beta} & \frac{\partial S_3}{\partial \gamma}
  \end{pmatrix}
= R  \arraycolsep=1.4pt
  \begin{pmatrix}
  Q_1 & Q_3 & -Q_2 \\
  -Q_3 & Q_2 & Q_1 \\
  Q_2 & -Q_1 & Q_3
  \end{pmatrix}\\
\label{eq:dSdt}  
\frac{\partial S}{\partial t} &=
  \begin{pmatrix}
  \frac{\partial S_1}{\partial t_1} & \frac{\partial S_1}{\partial t_2} & \frac{\partial S_1}{\partial t_3} \\
  \frac{\partial S_2}{\partial t_1} & \frac{\partial S_2}{\partial t_2} & \frac{\partial S_2}{\partial t_3} \\
  \frac{\partial S_3}{\partial t_1} & \frac{\partial S_3}{\partial t_2} & \frac{\partial S_3}{\partial t_3}
  \end{pmatrix}
 = I_{3\times3}   
\end{align}}

The derivation is straightforward. 

\subsection{Example: pinhole model with one radial distortion coefficient}

Here we use the pinhole model with one radial distortion coefficient as an example. 
It can be easily generalized to other more complicated models.
The model can be represented as:
\begin{eqnarray}
q_x(S) & = & u + (1 + k_1 r^2) f \frac{S_1}{S_3} \\
q_y(S) & = & v + (1 + k_1 r^2) f \frac{S_2}{S_3}
\end{eqnarray}
where $r = \sqrt{(\frac{S_1}{S_3})^2 + (\frac{S_2}{S_3})^2} = \frac{1}{S_3}\sqrt{S_1^2 + S_2^2}$. 
We define $\Theta = (f,u,v, k_1)$.

First, the partial derivatives of the local projection functions with respect to $S$ can be written as:
\begin{equation}
\begin{pmatrix}
\frac{\partial q_x}{\partial S} \\
\frac{\partial q_y}{\partial S}
\end{pmatrix}
=
\begin{pmatrix}
\frac{\partial q_x}{\partial S_1} & \frac{\partial q_x}{\partial S_2} & \frac{\partial q_x}{\partial S_3}\\
\frac{\partial q_y}{\partial S_1} & \frac{\partial q_y}{\partial S_2} & \frac{\partial q_y}{\partial S_3}\\
\end{pmatrix}
\end{equation}
where
\begin{align*}
    \frac{\partial q_x}{\partial S_1} &= \frac{f}{S_3} + \frac{f k_1}{S_3^3}( 3S_1^2 +S_2^2 ) \\
    \frac{\partial q_x}{\partial S_2} &= \frac{2 f k_1 S_1 S_2}{S_3^3} \\
    \frac{\partial q_x}{\partial S_2} &= -\frac{f S_1}{S_3^2} - \frac{3 f k_1 S_1 (S_1^2 + S_2^2)}{S_3^4}\\
    \frac{\partial q_y}{\partial S_1} &= \frac{2 f k_1 S_1 S_2}{S_3^3} \\
    \frac{\partial q_y}{\partial S_2} &= \frac{f}{S_3} +  \frac{f k_1}{S_3^3}(S_1^2 + 3S_2^2 ) \\
    \frac{\partial q_y}{\partial S_2} &= -\frac{f S_2}{S_3^2} - \frac{3 f k_1 S_2 (S_1^2 + S_2^2)}{S_3^4}\\
\end{align*}
% \begin{equation}
% \begin{pmatrix}
% \frac{1}{S_3} + \frac{k_1}{S_3^3}( 3S_1^2 +S_2^2 ) 
% & \frac{2 k_1 S_1 S_2}{S_3^3}
% & -\frac{S_1}{S_3^2} - \frac{3 k_1 S_1 (S_1^2 + S_2^2)}{S_3^4} \\
% \frac{2 k_1 S_1 S_2}{S_3^3}
% & \frac{1}{S_3} +  \frac{k_1}{S_3^3}(S_1^2 + 3S_2^2 )
% & -\frac{S_2}{S_3^2} - \frac{3 k_1 S_2 (S_1^2 + S_2^2)}{S_3^4} 
% \end{pmatrix}
% \end{equation}
This, together with Eq.~\eqref{eq:dSdR} and \eqref{eq:dSdt},
allows to compute the partial derivatives of residuals relative to extrinsic parameters in the Jacobian matrix $J$, by inserting into Eq.~\eqref{eq.pd1} and \eqref{eq.pd2} (and likewise for partial derivatives of $\hat{y}$).

As for the partial derivatives relative to intrinsic parameters, they can be computed as:
\begin{align}
\frac{\partial \hat{x}}{\partial \Theta} & =
\frac{\partial q_x}{\partial \Theta} = \begin{pmatrix} (1 + k_1 r^2)\frac{S_1}{S_3}
& 1 
& 0 
& r^2 f \frac{S_1}{S_3} 
\end{pmatrix} \\
\frac{\partial \hat{y}}{\partial \Theta} & =
\frac{\partial q_y}{\partial \Theta} = 
\begin{pmatrix} (1 + k_1r^2)\frac{S_2}{S_3}
& 0 
& 1 
& r^2 f\frac{S_2}{S_3} 
\end{pmatrix} 
\end{align} 
Now, we have everything needed for building $J$.

%%%%%%%%%%%%%%%%%%%%%%%%%%%%%%%%%%%%%%%%%%%%%%%%%%%%%%%%%%%%%%%%%
\section{Uncertainty map}
\label{sec:uncertainty-map}
Here we introduce the concept of uncertainty map which could be an effective tool to visualize the quality of the current calibration. Also, the uncertainty map can be used to evaluate the evolution of the quality along the calibration process when adding more and more images.

% \begin{figure*}[!ht]
%   \centering
%   \begin{tabular}{ccc}
% %      \vspace{-1em}
%     \includegraphics[width = 0.31\linewidth, trim = 5em 5em 2em 2.5em, clip]{img/umap_ini}  &
%     \includegraphics[width = 0.31\linewidth, trim = 5em 5em 2em 2.5em, clip]{img/umap_random} & 
%     \includegraphics[width = 0.31\linewidth, trim = 5em 5em 2em 2.5em, clip]{img/umap_wizard}  \\         
%     {\small (a)} &  {\small (b)} & {\small (c)} 
%   \end{tabular}
% \caption{Illustrations for the uncertainty maps and the effectiveness of our camera wizard. The size of the map is the same as acquired image size $640\times 480$. (a) Initial calibration results with 2 freely taken images. (b) Add two freely taken images. (c) Add two images proposed by wizard. We can clearly visualize: (1) the more calibration images, the lower the uncertainty (2) the uncertainty obtained from using wizard images is much lower than with freely taken ones (cf. the scale on the right hand side of each figure).}
% \label{fig:umap-illu}
% \end{figure*}
\begin{figure*}[!ht]
  \centering
  \begin{tabular}{cccc}
%      \vspace{-1em}
    \includegraphics[width = 0.28\linewidth, trim = 10em 10em 10em 2.5em, clip]{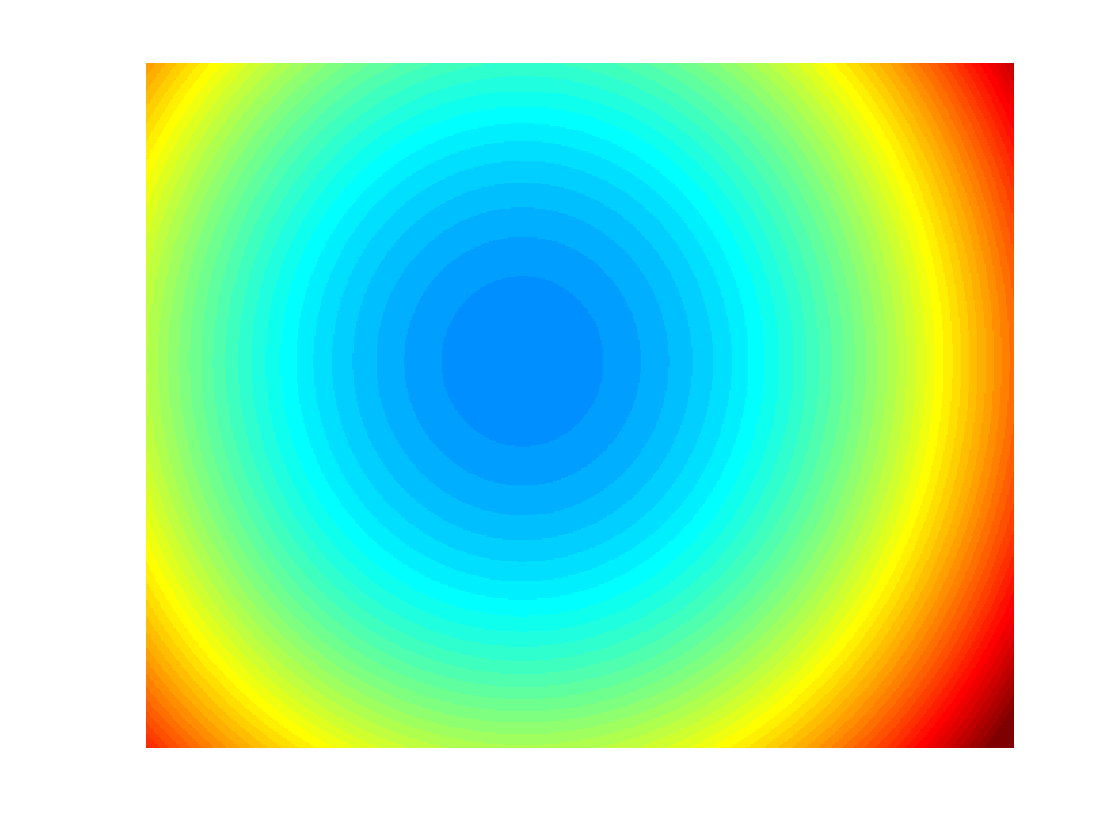}  &
    \includegraphics[width = 0.28\linewidth, trim = 10em 10em 10em 2.5em, clip]{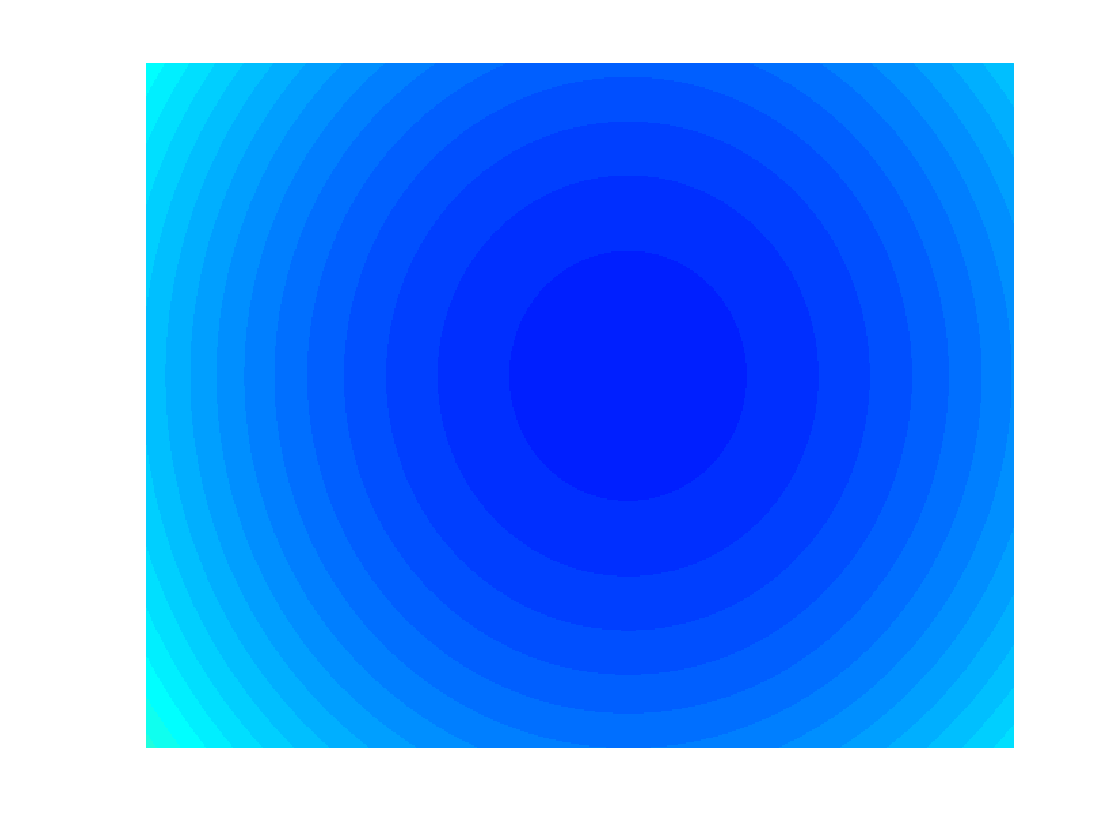} & 
    \includegraphics[width = 0.28\linewidth, trim = 10em 10em 10em 2.5em, clip]{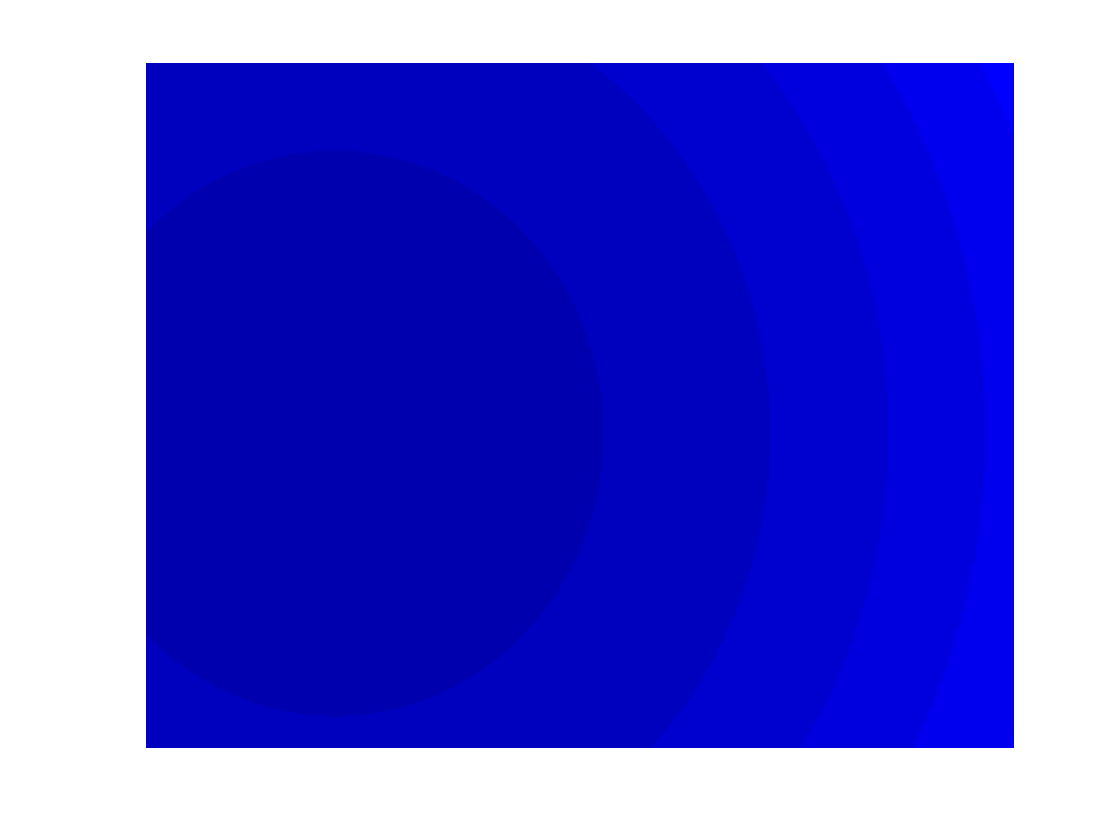} &
    \includegraphics[width = 0.041\linewidth]{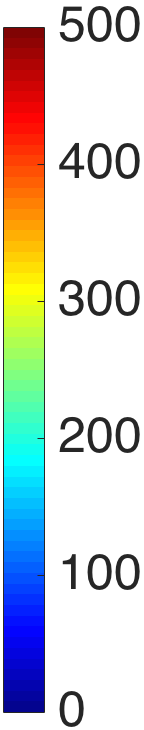} 
    \\     
    {\small (a)} &  {\small (b)} & {\small (c)} &
  \end{tabular}
\caption{Illustrations for the uncertainty maps and the effectiveness of our camera wizard. The size of the map is the same as acquired image size $640\times 480$. (a) Initial calibration results with 3 freely taken images. On top of the 3 freely-taken images, (b) add two freely taken images. (c) add two images proposed by wizard. We can clearly visualize: (1) the more calibration images, the lower the uncertainty (2) the uncertainty obtained from using wizard images is much lower than with freely taken ones (cf. the scale on the right hand side).}
\label{fig:umap-illu}
\end{figure*}

Given the $k \times k$ covariance matrix $\Sigma$ of intrinsic parameters, an uncertainty map is defined as the expected uncertainties of the local projection model across the image plane. 
That is to say, we propagate the uncertainties of intrinsic parameters to each pixel on the image area.
The quality of calibration at any stage of the calibration process can be visualized as shown in Fig.~\ref{fig:umap-illu}.

%In order to acquire the map, the following process should be performed.
Mathematically speaking, for each pixel $(x,y)$ on the image, the uncertainty propagation process can be formulated as:
\begin{equation}
\begin{split}
    \Gamma&(x,y) =\\& \begin{pmatrix}[1.5] \frac{\partial q_x(\Theta,S)}{\partial \Theta} |_{S=S(x,y,\Theta)} \\ \frac{\partial q_y(\Theta,S)}{\partial \Theta} |_{S=S(x,y,\Theta)} \end{pmatrix}	
    \Sigma
    \begin{pmatrix}[1.5] \frac{\partial q_x(\Theta,S)}{\partial \Theta}  |_{S=S(x,y,\Theta)} \\ \frac{\partial q_y(\Theta,S)}{\partial \Theta}  |_{S=S(x,y,\Theta)} \end{pmatrix}^\top	
    \label{eq:uncertainty-propergate}
\end{split}    
\end{equation}
where $\Gamma$ is a $2\times 2$ covariance matrix expressing the uncertainty per image point $(x,y)$.
To compute this, we first need to back-project the image point to a 3D point $S$ (any 3D point along the line of sight associated with the image point will do).
We denote this as $S(x,y,\Theta)$ in the above equation: the back-projection depends on the image point coordinates $x$ and $y$ and of course on the intrinsic parameters $\Theta$.
Then, we propagate the uncertainty on the intrinsic parameters (covariance matrix $\Sigma$) in a standard way through the forward projection, as shown in the above equation, to obtain the covariance matrix $\Gamma$.

Note that, we are aware that minimizing over this pixel reprojection uncertainty is better than over the trace of the covariance matrix $\Sigma$. The main problem with minimizing reprojection uncertainty is computation time, incompatible with real-time. In this paper we stuck to the trace, a commonly used simple cost function. In the future, we will consider the reprojection uncertainty for a subset of pixels. 

In the following, we provide an example for the process of acquiring the uncertainty map and some analysis.

\subsection{Example: 3-parameter pinhole model}

For this simple camera model, the uncertainty map can be analyzed theoretically, as shown in the following.
We remind the definition of the 3-parameter pinhole model, where the vector of intrinsic parameters is $\Theta=(f,u,v)^\top$.

\begin{eqnarray*}
q_x(\Theta,S) & = & u + f \frac{S_1}{S_3} \\
q_y(\Theta,S) & = & v + f \frac{S_2}{S_3}
\end{eqnarray*}

These are the ``forward'' projection equations.
As explained above, to construct the uncertainty map, we should back-project each pixel to 3D, i.e.\ to some 3D point $S$ which, if forward-projected, gives rise to the original pixel.
One possibility for the back-projection is as follows:
\[
S(x,y,\Theta) = \begin{pmatrix} x-u \\ y-v \\ f \end{pmatrix}
\]

We now compute the partial derivatives appearing in Eq.~\eqref{eq:uncertainty-propergate}:
\begin{eqnarray*}
\begin{pmatrix}
\frac{\partial q_x(\Theta,S)}{\partial \Theta} |_{S=S(x,y,\Theta)} \\
\frac{\partial q_y(\Theta,S)}{\partial \Theta} |_{S=S(x,y,\Theta)}
\end{pmatrix} & = &
\begin{pmatrix}
\frac{S_1}{S_3} & 1 & 0 \\
\frac{S_2}{S_3} & 0 & 1
\end{pmatrix} |_{S=S(x,y,\Theta)} \\
& = &
\begin{pmatrix}
\frac{x-u}{f} & 1 & 0 \\
\frac{y-v}{f} & 0 & 1
\end{pmatrix}
\end{eqnarray*}

Inserting this in Eq.~\eqref{eq:uncertainty-propergate}, we get the desired covariance matrices per image point:
\begin{equation*} 
\Gamma(x,y) =
\begin{pmatrix}
\frac{x-u}{f} & 1 & 0 \\
\frac{y-v}{f} & 0 & 1
\end{pmatrix}
\Sigma
\begin{pmatrix}
\frac{x-u}{f} & \frac{y-v}{f} \\
1 & 0 \\
0 & 1
\end{pmatrix}
\end{equation*}

To analyze the nature of the uncertainty map, we can imagine it as the coefficients of an ``uncertainty ellipse'' centered at every image pixel.
Once we acquire $\Gamma$ for each pixel, the trace, larger eigenvalue or determinant of $\Gamma$ can be used to measure the impact of the uncertainty of intrinsic parameters, on this pixel.
In Fig.~\ref{fig:umap-illu}, we generate the map by creating an image where the value of every pixel is given by the trace of $\Gamma$.

We may further analyze the nature of these uncertainty maps as follows.
When computing the trace of $\Gamma$ in detail, we get:
\begin{align*}
    tr(\Gamma(x,y)) = \Sigma_{22} + &\Sigma_{33} + \frac{\Sigma_{11}}{f^2} \left[ (x-u)^2 + (y-v)^2 \right]\\
&+ \frac{2}{f} \left[ \Sigma_{12} (x-u) + \Sigma_{13}(y-v) \right]
\end{align*}
This is quadratic with respect to $x$ and $y$. The uncertainty map, if visualized in 3D (as height map above the $x-y$-plane), can be shown to be a circular paraboloid.
The global minimum of the trace is attained at
\[
x^* = u - f \frac{\Sigma_{12}}{\Sigma_{11}}
\enspace\enspace\enspace
y^* = v - f \frac{\Sigma_{13}}{\Sigma_{11}}
\]
and has the following value:
\begin{equation*}
\Sigma_{22} + \Sigma_{33} - \frac{\Sigma_{12}^2 + \Sigma_{13}^2}{\Sigma_{11}}
\end{equation*}
A few observations are as follows. 
If the covariance between the focal length and the principal
point coordinates ($\Sigma_{12}$ and $\Sigma_{13}$) approaches zero, then the minimum of the uncertainty map coincides with the
principal point and its value there is the sum of the variances of the principal point coordinates.
And if the variance of the focal length tends to zero, the uncertainty map tends to being uniform.
Therefore, when the calibration result gets more accurate (low uncertainty), the range of the uncertainty map will become smaller, and its minimum will be approaching the principal point.
Finally, an empirical observation is that the covariance matrices $\Gamma$, when visualized as uncertainty ellipses, tend to be oriented towards the minimum of the uncertainty map (i.e.\ at each image point, the associated uncertainty ellipse has an axis that ``points'' towards the minimum of the uncertainty map).

More analysis, also for more complex camera models, is possible.

In summary, we believe that this concept is a principled way of displaying and interpreting the uncertainty of a calibration estimate.

%%%%%%%%%%%%%%%%%%%%%%%%%%%%%%%%%%%%%%%%%%%%%%%%%%%%%%%%%%%%%%%%%

%{\small
%\bibliographystyle{ieee}
%\bibliography{egbib}
%}
%
%
%
%\end{document}